\documentclass[twoside,11pt]{article}

\usepackage{blindtext}

\usepackage{jmlr2e}
\usepackage{amsmath}
\usepackage{graphics}
\usepackage{graphicx}
\usepackage{txfonts}
\usepackage{subfigure}
\usepackage{caption}
\usepackage{float}
\usepackage{threeparttable}
\usepackage{url}
\usepackage{multirow}
\usepackage{diagbox}
\usepackage{float}
\usepackage{booktabs}
\usepackage{color}
\usepackage[linesnumbered,ruled]{algorithm2e}

\usepackage{lastpage}
\jmlrheading{24}{2024}{1-\pageref{LastPage}}{9/23;}{}{21-0000}{Yiran Dong, Di Fan and Chuanhou Gao}

\ShortHeadings{LILI clustering algorithm: Limit Inferior Leaf Interval Integrated into Causal Forest for Causal Interference}{Yiran Dong, Di Fan and Chuanhou Gao}
\firstpageno{1}

\begin{document}

\title{LILI clustering algorithm: Limit Inferior Leaf Interval Integrated into Causal Forest for Causal Interference}

\author{\name Yiran Dong \email 12235045@zju.edu.cn \\
       \addr School of Mathematical Sciences\\
	   Zhejiang University\\
	   Hangzhou 310058, China.
       \AND
       \name Di Fan \email fandi@zju.edu.cn \\
       \addr School of Mathematical Sciences\\
	   Zhejiang University\\
	   Hangzhou 310058, China.
       \AND
       \name Chuanhou Gao \email gaochou@zju.edu.cn \\
       \addr School of Mathematical Sciences \\
	  Zhejiang University\\
       Hangzhou 310058, China.}

\editor{My editor}

\maketitle

\begin{abstract}
Causal forest methods are powerful tools in causal inference. Similar to traditional random forest in machine learning, causal forest independently considers each causal tree. However, this independence consideration increases the likelihood that classification errors in one tree are repeated in others, potentially leading to significant bias in causal effect estimation. In this paper, we propose a novel approach that establishes connections between causal trees through the Limit Inferior Leaf Interval (LILI) clustering algorithm. LILIs are constructed based on the leaves of all causal trees, emphasizing the similarity of dataset confounders. When two instances with different treatments are grouped into the same leaf across a sufficient number of causal trees, they are treated as counterfactual outcomes of each other. Through this clustering mechanism, LILI clustering reduces bias present in traditional causal tree methods and enhances the prediction accuracy for the average treatment effect (ATE). By integrating LILIs into a causal forest, we develop an efficient causal inference method. Moreover, we explore several key properties of LILI by relating it to the concepts of limit inferior and limit superior in the set theory. Theoretical analysis rigorously proves the convergence of the estimated ATE using LILI clustering. Empirically, extensive comparative experiments demonstrate the superior performance of LILI clustering.
\end{abstract}

\begin{keywords}
  Causal Inference, Causal Forest, Limit Inferior Leaf Interval, Average Treatment Effect, Zero-one Law
\end{keywords}

\section{Introduction}
Causation in statistics focuses on identifying and analyzing causal relationships among variables (\cite{pearl2003statistics}). Causal inference extends this framework by incorporating advanced machine learning techniques to facilitate a deeper understanding of datasets, enabling the estimation of causal effects and providing insights into the underlying relationships. This capability is crucial for guiding decision-making and formulating effective policies in practical applications.
With the increasing use of black-box models such as neural networks, interpretability of causal inference has become more critical. Causal inference also offers a practical alternative to expensive A/B tests, such as randomized controlled trials in drug and disease studies (\cite{thomas2020using, shi2022learning, prosperi2020causal}) or sociological policy experiments (\cite{imbens2015causal, angrist2009mostly}). By using historical datasets, causal inference can effectively simulate A/B tests. In the domain of recommendation systems, it improves metrics like click-through rate and conversion rate by addressing biases (\cite{chen2023bias, wang2021clicks}), estimating uplift (\cite{bonner2018causal}), and explaining causal relationships (\cite{xu2023causal}). As machine learning continues to progress, causal inference methods are becoming more powerful, with applications expanding across various fields.

Unlike traditional machine learning tasks, causal inference divides instances into treatment and control groups based on the treatment variables in the dataset and estimates the difference in their outcomes, known as the average treatment effect (ATE). Causal inference methods can be broadly categorized into four main classes: re-weighting methods, tree-based models, matching methods, and meta-learning models. 
In the re-weighting domain, inverse propensity weighting (IPW) (\cite{rosenbaum1987model, rosenbaum1983central}) adjusts instance weights based on the conditional distribution of treatment variables, enabling direct computation of the ATE. Doubly robust estimator (\cite{robins1994estimation}) combines two regression models trained on the treatment and control groups and adjusts instance weights using these models within the IPW framework. Data-driven variable decomposition (\cite{kuang2017treatment}) separates features into confounders and adjustments by solving an optimization problem and applies distinct estimators to generate weights for each group. 
Uplift forest (\cite{rzepakowski2012decision}), a classic tree-based model, adapts the decision tree framework for causal inference tasks. It considers instances within the same leaf node but subjected to different treatments as counterfactual outcomes for one another. Unlike traditional random forests, uplift forest employs Kullback-Leibler (KL) divergence or $\chi^2$-divergence as splitting criteria. \cite{hill2011bayesian} introduces the application of Bayesian additive regression trees (BART) (\cite{chipman2010bart}) for causal inference, leveraging the natural advantages of BART, such as its ability to statistically model and identify heterogeneous treatment effects.
Matching algorithms estimate counterfactual outcomes by calculating the distances between individuals and other instances, using these distances and corresponding observed outcomes. For example, randomized nearest neighbor matching (RNNM) (\cite{li2016matching}) projects the original feature space into a lower-dimensional space using random projections and subsequently identifies the top-k nearest neighbors for each instance. Similarly, the coarsened exact matching (\cite{iacus2012causal}) algorithm employs a specialized coarsening technique to group instances into discrete categories and calculates weights for target samples based on the proportion of treatment and control samples within the same category.
Meta-learning models utilize advanced predictive frameworks, such as deep neural networks, to address causal inference tasks. Two-learner (T-learner) (\cite{kunzel2019metalearners}) trains separate neural networks for treatment and control groups. For a given instance, T-learner predicts the outcomes under both treatment and control conditions and calculates the treatment effect as the difference between these predictions. Similarly, the causal effect variational autoencoder (\cite{louizos2017causal}) employs a variational autoencoder framework to model the conditional distributions of outcomes for treatment and control groups.

Among the four categories of causal inference methods, tree-based models are particularly popular due to their efficiency and accuracy in estimating the ATE. Consequently, numerous studies have focused on optimizing tree-based models (\cite{green2012modeling, foster2011subgroup}). \cite{hill2013assessing} propose a new approach to identify the common causal variables to optimize BART models in high-dimensional space. Causal forest (\cite{wager2018estimation}) introduces specific assumptions and rules to enhance the construction and estimation of each causal tree. We focus on causal forest due to its superior performance, and a detailed introduction to causal forest is provided in the next section.
It is important to note that trees operate independently within all the tree-based methods mentioned before. However, for each leaf node in causal trees, there is a high likelihood that instances are mistakenly classified. This mistake introduces bias into the estimation of ATE, and simply adding more causal trees does not inherently resolve this issue because the trees remain independent. Therefore, finding a way to connect independent causal trees and exclude mistakenly classified instances becomes a critical aspect of optimizing tree-based models.

To address this, we propose constructing a new cluster for each instance by aggregating all the leaf nodes that contain the instance across the causal trees. If another instance consistently appears in the same leaves as the given instance across a sufficient number of trees, we classify them into the same cluster, indicating a high level of similarity. This clustering method resembles the limit inferior operation in set theory, which we refer to as limit inferior leaf interval (LILI). In addition, the main contributions of this paper are as follows:
\begin{itemize}
    \item We establish a connection between LILI and the traditional concepts of limit inferior and limit superior in set theory;
    \item  We prove that LILI possesses a key advantageous property absent in ordinary limit inferior operations: the zero-one law (\cite{billingsley2017probability}). Using the zero-one law, LILI reduces bias and enhances the accuracy of estimated ATE;
    \item The convergence of the estimated distribution induced by LILI requires fewer and less stringent assumptions compared to other tree-based models;
    \item  Fine-tuning methods for hyperparameters and specific function are provided.
\end{itemize}
As a result, the estimated ATE by the proposed algorithm converges more efficiently. 

The rest of this paper is organized as follows: Section 2 provides a detailed introduction about causal inference, the construction of causal trees, and the prediction process of causal forest. In Section 3, we define LILI and demonstrate its properties by linking it to limit superior and limit inferior. Section 4 presents the convergence proof of the estimated ATE and extends the LILI framework. The experimental results and hyperparameter analyses are given in Section 5. Finally, we conclude the paper in Section 6.

\textbf{Notation}: Uppercase letters are used to represent variables, while bold uppercase letters denote multi-dimensional variables, such as $T, \mathbf X, Y$. Variables with subscript indices, like $T_i, \mathbf X_i, Y_i$, or lowercase letters, $t, \mathbf x, y$, refer to the values of the $i$th instance or some random values of corresponding variables. The standard notation for bold letters, such as $X_i$, represents the $i$th feature of the vector $\mathbf X$. And decorative uppercase letters indicate sets. The parameters or hyperparameters of models are represented by Greeks. Otherwise, some specific symbols may be redefined within the context as needed.

\section{Preliminaries}
In this section, we first introduce the fundamental concepts of causal inference. Then, we outline the principles of causal forest and their approach to predicting the ATE in causal inference tasks.

Features in causal inference datasets can be categorized into three components $(T, \mathbf X, Y)$. Here, $T$ is the treatment variable, $\mathbf X=(X_1, X_2, ..., X_d)$ are covariates that contain all other relevant variables, and $Y$ is the label. Generally, $T_i=1$ indicates that unit $i$ is part of the treatment group, while $T_i=0$ signifies membership in the control group. Although real-world experiments often involve multiple treatment measures and there is a growing body of research on multiple treatments (\cite{vanderweele2013causal, scotina2019matching, li2019propensity}), we concentrate only on the binary treatment variable. Different from traditional machine learning tasks that primarily aim to predict $Y$, causal inference emphasizes understanding the individual treatment effect (ITE), defined as $ITE(i) = Y_i(T=1)-Y_i(T=0)$ where $Y_i(1)$ (short for $Y_i(T=1)$) and $Y_i(0)$ (short for $Y_i(T=0)$) are potential outcomes for unit $i$ in the treatment and control groups, respectively. However, we can only observe one outcome, $Y_i(1)$ or $Y_i(0)$, for each unit in the dataset. In other words, unit $i$ cannot simultaneously belong to both the treatment and control groups. If we observe $T_i = 1$ in the data, then $Y_i(0)$ is treated as the counterfactual outcome for unit $i$. The primary objective of causal inference is to predict these counterfactual outcomes for all units and estimate the average treatment effect (ATE):
\begin{equation}
\begin{aligned}
ATE &= E[Y_i(1)-Y_i(0)]\\
    &= E_{\mathbf X}[Y(1)|\mathbf X = \mathbf X_i] - E_{\mathbf X}[Y(0)|\mathbf X = \mathbf X_i]\\
    &= E_{\mathbf X}[Y|T=1, \mathbf X = \mathbf X_i]-E_{\mathbf X}[Y|T=0, \mathbf X = \mathbf X_i].
\end{aligned}
\end{equation}

To simplify the causal inference tasks, the following three commonly-used assumptions are needed (\cite{yao2021survey}).

\begin{itemize}
    \item \textit{Stable Unit Treatment Value Assumption (SUTVA)}: Units are independent of each other and there are no interactions between units. The potential outcome of one unit may not affect the potential outcome of others;
    \item \textit{Ignorability}: Given all covariates, the treatment is independent of potential outcomes, i.e. $T\perp\!\!\!\perp Y(0), Y(1)\ |\ \mathbf X$;
    \item \textit{Positivity}: The assignment of treatment is not deterministic given any value of covariates, i.e., $P(T|\mathbf X=\mathbf X_i) > 0$ for any $\mathbf X_i$.
\end{itemize}
The \textit{SUTVA} assumption eliminates randomness arising from the interactions among units, enabling a focus on the relationships between features rather than units. The \textit{Ignorability} assumption consists of two key components. First, for any given unit, the potential outcomes should not influence the treatment assignment. Secondly, it states that $P(Y(1), Y(0)|\mathbf X=\mathbf X_i, T=1)=P(Y(1), Y(0)|\mathbf X=\mathbf X_i, T=0)$, indicating that the distribution of potential outcomes cannot be affected by the actual treatment assignment. If $T$ can be determined for some $\mathbf X=\mathbf x_j$, then the counterfactual of $\mathbf x_j$ will remain unobserved, making its prediction impossible and meaningless. Consequently, the \textit{positivity} assumption is essential. Together, these three assumptions make the problem well-defined and solvable. 

Building on the aforementioned assumptions, tree-based causal inference methods use tree models for causal tasks. Similar to the random forest, a causal forest consists of multiple trees, known as causal trees. The construction of a causal tree closely resembles that of a traditional decision tree in machine learning. It employs a similar scoring mechanism to select split features and values at each node, with growth termination governed by specific criteria. However, causal trees conceptualize the tree model not just as a prediction model but also as a clustering algorithm, similar to k-nearest neighbors. In this framework, instances within the same leaf form a cluster, but with the structural integrity of a decision tree. 
As a result, causal forests differ in their choice of split features, split values, and stopping rules  (\cite{wager2018estimation}).

\begin{itemize}
    \item \textit{Data sampling}: The dataset for constructing a single tree with size $s_n$ is obtained by subsampling from the entire set of $n$ instances. As $n$ increases, $s_n$ also approaches infinity. 
    \item \textit{Honesty}: The dataset for the construction of each causal tree is split into two parts. One part is used for deciding the split features and split values, while the other part is used for estimating ATE.
    \item \textit{$\alpha$-regularity}: For a node that is not a leaf, its child contains at least $\alpha s_n$ instances where $0<\alpha\leq 0.2$. Moreover, the leaf node contains the data with a size between $l$ and $2l-1$ for some integer $l$.
    \item \textit{Random-split}: The probability that every feature is chosen as a split feature at every split step is bounded below by $\pi/d$ for some $0<\pi<1$. 
\end{itemize}

By following the rules outlined above, causal forest treats the instances within the same leaf with different treatment assignments as counterfactual instances of each other. For an instance $\mathbf X = \mathbf{x}$, the individual treatment effect (ITE) predicted by the $k$th tree can be calculated as:
\begin{align}
    \widehat{ITE}_k(\mathbf x) = \frac{1}{\left |L^t_k(\mathbf x)\right |}\sum_{i\in L^t_k(\mathbf x)} Y_i - \frac{1}{\left |L^c_k(\mathbf x)\right |}\sum_{i\in L^c_k(\mathbf x)}Y_i,
\end{align}
where $L_k(\mathbf x)$ is the leaf that contains instance $\mathbf x$ in $k$th tree, and $L^t_k(\mathbf x) \left (L^c_k(\mathbf x) \right )$ is the instance of the treatment group (control group) in $L_k(\mathbf x)$, i.e. $\{\mathbf X_i\in L_k(\mathbf x), T_i=1\} \left( \{\mathbf X_i\in L_k(\mathbf x), T_i=0\}\right)$. Note that the causal tree exclusively selects split features on covariates, with each value of $\mathbf X$ being assigned to only one leaf node. Even if there are two instances that share the same covariates but have different treatments, they will inevitably fall within the same leaf. This eliminates any potential confusion in the representation $L(\mathbf x)$. Thus, the final ITE of $\mathbf x$ is obtained by $\widehat{ITE}(\mathbf x) = \frac1K\sum^K_{k=1} \widehat{ITE}_k(\mathbf x)$. By traversing all the instances in the dataset, we obtain the ATE for the causal forest.

Although every causal tree is trained independently, they all rely on the same dataset and features. Given that the tree models are trained under the greedy strategy, treating instances in the same leaf as counterfactual counterparts may not be entirely justified. However, the causal forest attempts to mitigate the effects of this greedy approach by incorporating randomness through measures such as \textit{data sampling} and \textit{random-split}. Ultimately, it simply averages the ITEs of causal trees and takes the mean value as its outcomes, which may overlook the interdependencies between the trees.
To address these limitations, we propose optimized methods in the following section.

\section{Limit Inferior Leaf Interval}
In this section, we first introduce the definition of the LILI and analyze the biases inherent in the original causal forest methodology, explaining how LILI effectively mitigate these biases. To establish a clearer correspondence between LILI and the conventional concepts of limit inferior, we define the Limit Superior Leaf Interval (LSLI). Furthermore, we highlight certain characteristics of LILI and LSLI, such as the zero-one law, which is absent in conventional limit inferior. The symmetry between LILI and LSLI is further explored. Finally, we provide a detailed discussion of the LILI clustering algorithm and its implications.

Each split step in the causal tree only considers one feature $X_i$ and a single split value $a$, classifying the instance $\mathbf x$ based on whether $x_i\leq a$ or $x_i>a$. Consequently, the leaf node, as the terminal node of the causal tree, can be represented by the Borel set $\prod_{i=1}^{d} (a_i, b_i)$ where $a_i, b_i\in (-\infty, +\infty)$ are the lower and upper bound of split values for the $i$th feature. We denote the leaf node interval to which the instance $\mathbf X=\mathbf x$ belongs in the $k$th causal tree trained on dataset $\mathcal D$ as $I(\mathbf x, \theta_k, \mathcal D)$, where $\theta_k$ comprises all split features and values of the $k$th tree. Thus, we can decide a tree given a $\theta$.
Gathering all the split features and values,  $\Theta_K = \{\theta_1, ..., \theta_K\}$ represents a forest with $K$ causal trees. Unlike $L_k(\mathbf x)$ in \textbf{Section 2}, $I(\mathbf x, \theta_k, \mathcal D)$ indicates the interval defined by each split step, whereas $L_k(\mathbf x)$ is the set of instances in the leaf. Using these intervals, we give the main definition of this study.

\begin{definition}
\label{LILI}
For the instance $\mathbf X = \mathbf x$, we define the limit inferior leaf interval with $K$ trees under $\mathcal D$ as 
\begin{align}
    I(\mathbf x, \Theta_K, \mathcal D) = \bigcup_{1\leq i_1<\cdots i_s\leq K\atop s> K-\sqrt{K}}\bigcap_{k=1}^{s} I(\mathbf x, \theta_{i_k}, \mathcal D),
\end{align}
where $K$ is the total number of distinct causal trees contained in the causal forest, and $\theta_{i_k}$ are all distinct. According to this, we define the limit inferior leaf interval (LILI) as 
\begin{align}
    I(\mathbf x) = \lim_{K\to\infty}\lim_{n\to\infty} I(\mathbf x, \Theta_K, \mathcal D),
\end{align}
where $n$ is the total number of instances.
\end{definition}

The LILI must be constructed with infinite instances. Otherwise, the total number of possible causal trees becomes limited. When the number of instances is finite, the split values and features are also limited, resulting in only a finite variety of causal trees, despite potentially infinite repetitions of the same trees. Consequently, analyzing the connections among identical causal trees becomes meaningless. Therefore, it is essential for $n\to\infty$, and the sequence of $K\to\infty$ and $n\to\infty$ cannot be interchanged. For simplicity and clarity, we omit the $\mathcal D$ and write $I(\mathbf x, \theta_i, \mathcal D)$, $I(\mathbf x, \Theta_K, \mathcal D)$ as $I(\mathbf x, \theta_i)$, $I(\mathbf x, \Theta_K)$ in the rest of the paper.

Let us examine the precise meaning of $I(\mathbf x, \Theta_K)$. If another instance $\mathbf y\neq \mathbf x$ belongs to $I(\mathbf x)$, then there will be a sequence of leaves $j_1, j_2, ..., j_s$ where $s\geq K-\sqrt{K}$ such that $\mathbf y \in \bigcap_{k=1}^s I(\mathbf x, \theta_{j_k})$, i.e., $\mathbf y$ belongs to all leaf nodes in these trees. Therefore, $\mathbf y \in I(\mathbf x)$ if and only if there are at least $K-[\sqrt{K}]+1$ trees that assign $\mathbf x$ and $\mathbf y$ in the same leaves where $[\sqrt{K}]$ is the integer part of $\sqrt{K}$.

Ultimately, LILI provides a robust method for classifying instances into distinct groups. Unlike traditional approaches that merely average outcomes across all leaves, LILI seeks to integrate the classification results from all trees. This brings attention to a potential bias inherent in causal forests.

\begin{lemma}
\label{lb-cf}
Suppose that $\mathbf X=\mathbf x, \mathbf y$ are two different instances, and one of their features has the same value, $x_1 = y_1$. If the construction of causal trees follows the four rules in \textbf{Section 2}, then given the number of instances, the probability that $\mathbf x$ and $\mathbf y$ are assigned to the same leaf has a lower boundary
\begin{align}
\label{lower-bound}
    \left(\frac{\pi}{d}\right)^{\left\lceil\frac{\ln\alpha}{\ln(1-\alpha)}\right\rceil} \leq P\left(\mathbf y\in L(\mathbf x)\mid\mathcal D\right),
\end{align}
where $\left\lceil\frac{\ln\alpha}{\ln(1-\alpha)}\right\rceil$ is the round-up of $\frac{\ln\alpha}{\ln(1-\alpha)}$.
\end{lemma}

The detailed proofs of all lemmas, propositions, and theorems appear in the Appendix. In this lemma, the condition $x_1=y_1$ posits that feature $X_1$ cannot be utilized to split instances $\mathbf x$ and $\mathbf y$. Simultaneously, $\mathbf x, \mathbf y$ differ significantly across all other features, rendering them incapable of serving as counterfactual outcomes for each other. In high-dimensional datasets, it is common for instances to share identical values on less relevant or discrete features while exhibiting substantial differences on critical features. In this situation, $\mathbf x, \mathbf y$ cannot act as counterfactual outcomes. However, the causal forest under the properties in \textbf{Section 2} maintains a lower bound for $P\left(\mathbf y\in L(\mathbf x)\mid\mathcal D\right)$. Notably, the left side of equation (\ref{lower-bound}) is invariant to $n$, indicating that irrespective of the dataset size, there exists at least $(\pi/d)^{\left\lceil\frac{\ln\alpha}{\ln(1-\alpha)}\right\rceil}$ likelihood that causal trees erroneously assign $\mathbf x, \mathbf y$ to the same leaf. As the number of similar or redundant features increases, this lower bound also rises linearly. This bias can affect the final estimated ATE and cannot be mitigated by simply increasing the number of causal trees, as each tree operates independently and may carry this bias. In contrast, we will demonstrate that such bias is unlikely to arise within the LILI framework in the rest of this section.

We call $I(\mathbf x)$ the limit inferior not just because it is similar to the definition of the limit inferior of the sequence of sets $\{A_1, A_2, ...\}$, $\liminf_{i\to\infty} A_i =  \bigcup_{m=1}^\infty\bigcap_{i=m}^\infty A_i$, but also because it has many same properties as limit inferior.

\begin{proposition}
\label{LSLI}
We denote the complementary set of $I(\mathbf x, \theta_k)$ as $I^c(\mathbf x, \theta_k)$, then 
\begin{align*}
    I^c(\mathbf x, \Theta_K) = \left(\bigcup_{1\leq i_1<\cdots i_s\leq K\atop s> K-\sqrt{K}}\bigcap_{k=1}^{s} I(\mathbf x, \theta_{i_k})\right)^c = \bigcap_{1\leq i_1<\cdots i_s\leq K\atop s>K-\sqrt{K}}\bigcup_{k=1}^{s} I^c(\mathbf x, \theta_{i_k}).
\end{align*}
The above equation also holds when $K, n\to\infty$. According to this, we define the limit superior leaf interval (LSLI) with $K$ trees and the limit superior leaf interval as 
\begin{align*}
\bar I(\mathbf x, \Theta_K) = \underset{1\leq i_1<\cdots i_s\leq K\atop s\leq K-\sqrt{K}}{\bigcap}\bigcup_{k=1}^{s} I(\mathbf x, \theta_{i_k}), \\
\bar I(\mathbf x) = \lim_{K\to\infty}\lim_{n\to\infty} \bar I(\mathbf x, \Theta_K).
\end{align*}
\end{proposition}

By \textbf{Proposition \ref{LSLI}}, we observe that the complementary set of LILI exhibits a structure closely resembling that of the limit superior of sets. $\limsup_{i\to\infty} A_i = \bigcap_{m=1}^\infty\bigcup_{i=m}^\infty A_i$. Considering the limit inferior and limit superior of sets follows
\begin{align*}
\left(\liminf_{i\to\infty} A_i\right)^c = \limsup_{i\to\infty} A_i^c,
\end{align*}
we use similar properties to induce the definition of LSLI. By establishing a connection between LILI and LSLI with the concepts of limit inferior and limit superior of sets, we obtain

\begin{proposition}
\label{LILI-LSLI}
For instance $\mathbf X = \mathbf x$, let $I(\mathbf x, \Theta_K)$ and $\bar I(\mathbf x, \Theta_K)$ be the LILI and LSLI with $K$ trees on $\mathbf x$, 
    \begin{itemize}
        \item[(1)] $I(\mathbf x, K)\subseteq \bar I(\mathbf x, K)$ and $I(\mathbf x)\subseteq \bar I(\mathbf x)$;
        \item[(2)] If $I(\mathbf x, \theta_1)\subseteq I(\mathbf x, \theta_2)\subseteq\cdots$, then 
        \begin{align*}
            I(\mathbf x) = \bar I(\mathbf x) = \bigcup^\infty_{k=1} I(\mathbf x, \theta_k);
        \end{align*}
        \item[(3)] If $I(\mathbf x, \theta_1)\supseteq I(\mathbf x, \theta_2)\supseteq\cdots $, then 
        \begin{align*}
            I(\mathbf x) = \bar I(\mathbf x) = \bigcap^\infty_{k=1} I(\mathbf x, \theta_k);
        \end{align*}
    \end{itemize}
\end{proposition}

Combining \textbf{Proposition \ref{LSLI}} and \textbf{Proposition \ref{LILI-LSLI}}, LILI and LSLI satisfy all the fundamental properties of limit inferior and limit superior of sets. Note that the conventional definitions of limit inferior and limit superior cannot be directly applied to causal trees, as they are only meaningful when there are already infinite causal trees. For the finite $K$ causal trees, the limit inferior becomes $\bigcup_{m=1}^K\bigcap_{k=m}^K I(\mathbf x, \theta_k) = I(\mathbf x, \theta_K)$,  which is not useful for building clusters.
Thus, it is reasonable to refer to $I(\mathbf x)$ as a limit inferior leaf interval, allowing us to draw deeper insights based on these properties.

\begin{theorem}
\label{core-lemma}
If the constructions of every causal tree are independent, then for any two instances $\mathbf X=\mathbf x, \mathbf y$, they are in the same LILI if and only if the probability that they are grouped into the same leaf converges to $1$, i.e.
\begin{align}
\label{LI-equivalence}
    P(\mathbf y\in I(\mathbf x)\mid\mathcal D)=1\Longleftrightarrow\lim_{n\to\infty} P\left(\mathbf y\in L(\mathbf x)\mid\mathcal D\right) = 1.
\end{align}
\end{theorem}

\textbf{Theorem \ref{core-lemma}} requires the independence of causal trees, i.e., the selection of split nodes and split values on one tree is not influenced by others. Under the rules in \textbf{Section 2}, the split nodes and split values depend on the sample instances and some randomness, while $\mathbf y\in L_k(\mathbf x)$ relies on the same dataset for all $k$. When given $\mathcal D$, the only differences among the causal trees arise from the methods of sampling and the selection of split features, which are independent of one another.
\begin{align*}
P(\mathbf y\in L_j(\mathbf x),\mathbf y\in L_k(\mathbf x)\mid\mathcal D) = P(\mathbf y\in L_j(\mathbf x)\mid\mathcal D)P(\mathbf y\in L_k(\mathbf x)\mid\mathcal D)
\end{align*}
with $j\neq k$. Therefore, equivalence \ref{LI-equivalence} holds only under the conditional probability given $\mathcal D$. This independence provides us with a strong condition to study LILI and is consistently achieved under the construction rules. We apply this independence throughout the rest of the paper without further mention and use $P(\mathbf y\in I(\mathbf x))$ and $P(\mathbf y\in L(\mathbf x))$ to represent $P(\mathbf y\in I(\mathbf x)\mid\mathcal D)$ and $P(\mathbf y\in L(\mathbf x)\mid\mathcal D)$ for short.

As we fixed the instances $\mathbf x, \mathbf y$, the randomness among event $\mathbf y\in I(\mathbf x)$ is how to construct infinity trees with given infinity dataset $\mathcal D_\infty$, i.e. $\{\Theta_\infty(\mathcal D_\infty)\}$ is the domain of the event $\mathbf y\in I(\mathbf x)$ where $\mathcal D_\infty$ is any possible infinite dataset and $\Theta_\infty(\mathcal D_\infty)$ is the parameters of infinity causal trees under $\mathcal D_\infty$. The expression $P(\mathbf y\in I(\mathbf x))=1$ signifies that for almost any $\Theta_\infty(\mathcal D_\infty)$, the event $\mathbf y\in I(\mathbf x)$ achieves, and it may fail on the subset of $\{\Theta_\infty(\mathcal D_\infty)\}$ with zero measure under the probability measure $P$.

Recall the situation in \textbf{Lemma \ref{lb-cf}}, where $\mathbf x,\mathbf y$ are similar or identical on a specific feature, such as $X_1$. However, they exhibit significant differences in other features, allowing any split feature aside from $X_1$ to divide $\mathbf x,\mathbf y$ into different nodes. It means that if $\mathbf y\in L(\mathbf x)$, every split feature of the node that contains $\mathbf x$ cannot be $X_1$, including the first split node. The probability that the first split feature is $X_1$ is bounded above by $1-\pi(d-1)/d$ according to \textit{Random-split} rules. Since every causal tree must at least have one split node, this upper bound also applies to $\mathbf y\in L(\mathbf x)$,
\begin{align*}
    P(\mathbf y\in L(\mathbf x))\leq 1-\frac{\pi(d-1)}{d}.
\end{align*}
Combining \textbf{Theorem \ref{core-lemma}}, they are unlikely to belong to the same LILI for sufficiently large $n$ and $K$, as the majority of causal trees are expected to separate them through splitting.

Therefore, LILI effectively eliminates this type of bias present in the original causal forest. This justification underpins our restriction of $s>K-\sqrt{K}$ in the definition of LILI. As $n,K$ approach infinity, the proportion of causal trees in which $\mathbf x, \mathbf y$ are not in the same leaves goes to zero, i.e. $\sqrt{K}/K\to 0$. Conversely, if we broaden the restriction of $s$, for example, $s>0.8K$, then $P\left(\mathbf y\in L(\mathbf x)\right)$ will obtain a lower bound greater than $0$, regardless of the dataset size or the extent of the causal forest.

\textbf{Theorem \ref{core-lemma}} reframes the inquiry from finding instances such that $P(\mathbf y\in L(\mathbf x))=1$ to finding instances for which $P(\mathbf y\in I(\mathbf x))=1$. However, this transformation does not necessarily simplify the problem, as we cannot guarantee that any instance $\mathbf y\in I(\mathbf x)$ will satisfy $P(\mathbf y\in I(\mathbf x))=1$. 

\begin{corollary}
\label{LILI-01}
If $\lim_{n\to\infty} P(\mathbf y\in L(\mathbf x))<1$, then $P(\mathbf y\in I(\mathbf x))=0$. Otherwise, if $\lim_{n\to\infty}P(\mathbf y\in L(\mathbf x))=1$, then $P(\mathbf y\in I(\mathbf x))=1$
\end{corollary}

\textbf{Corollary \ref{LILI-01}} can be directly inferred from the proof of \textbf{Theorem \ref{core-lemma}} and reveals a significant aspect of the LILI framework: the zero-one law. This law, an important property in set theory, typically applies to limit superior but may not hold for limit inferior. This corollary establishes that the zero-one law is indeed valid for LILI. Consequently, for an instance $\mathbf y\in I(\mathbf x)$ in practice, we can basically confirm $P(\mathbf y\in I(\mathbf x))=1$, as only $I(\mathbf x)$ in a null set of $\{\Theta_\infty(\mathcal D_\infty)\}$ group $\mathbf x, \mathbf y$ into the same leaf. Combining \textbf{Theorem \ref{core-lemma}} and \textbf{Corollary \ref{LILI-01}}, 
\begin{align*}
    \mathbf y\in I(\mathbf x)\quad\text{basically equivalent to}\quad P(\mathbf y\in I(\mathbf x)) = 1 \Longleftrightarrow \lim_{n\to\infty}P(\mathbf y\in L(\mathbf x))=1.
\end{align*}
Hence, we identify a method to collect instances with $P(\mathbf y\in L(\mathbf x))=1$ from $\mathbf y\in I(\mathbf x)$, which indicates LILI can effectively eliminate the bias in \textbf{Lemma \ref{lb-cf}}, as well as other potential sources of bias.

Similar to the causal forest and LILI, LSLI, as the limit superior of leaf nodes, offers an alternative approach to linking causal trees. However, LSLI is not well-suited for use as a clustering method. In fact, using LSLI could introduce even more bias than the original causal forest, making it less effective for clustering instances and estimating ATE.


\begin{theorem}
\label{LSLI-01}
If $\lim_{n\to\infty}P(\mathbf y\in L(\mathbf x))=0$, then $P(\mathbf y\in \bar I(\mathbf x))=0$. Otherwise, if $P(\mathbf y\in L(\mathbf x))>0$, then $P(\mathbf y\in \bar I(\mathbf x))=1$.
\end{theorem}

\textbf{Theorem \ref{LSLI-01}} establishes the zero-one law for LSLI. In contrast to the zero-one law on LILI, this law sets $P(\mathbf y\in L(\mathbf x))=0$ as the threshold for LSLI. Then we can directly derive the following conclusion:
\begin{align}
\label{LS-equivalence}
    P(\mathbf y\in \bar I(\mathbf x))=0 \Longleftrightarrow \lim_{n\to\infty}P(\mathbf y\in L(\mathbf x))=0 .
\end{align}
The equivalence (\ref{LS-equivalence}) and \textbf{Theorem \ref{core-lemma}} together highlight a profound symmetry between LSLI and LILI, and their relationships with $P(\mathbf y\in L(\mathbf x))$ are fundamentally opposite. Furthermore, $\mathbf y\in \bar I(\mathbf x)$ occurs in almost any $\Theta_\infty$ as long as $P(\mathbf y\in L(\mathbf x))>0$ for sufficiently large $n$, which is much easier to achieve than \textbf{Theorem \ref{core-lemma}}. However, in the scenarios described in \textbf{Lemma \ref{lb-cf}}, $\mathbf x, \mathbf y$ are invariably classified within the same LSLI, as $P(\mathbf y\in L(\mathbf x))$ maintains a lower bound greater than zero. In contrast, causal forest averages the outcomes across all causal trees, with the majority of them splitting $\mathbf x,\mathbf y$ away. Therefore, if we treat $\bar I(\mathbf x)$ as a cluster for computing ATE, this will not only fail to eliminate bias but may also exacerbate it.

A critical reason the limit superior $\limsup_{i\to\infty} A_i$ adheres to the zero-one law is that it belongs to the tail $\sigma$-field $\mathcal A$ which can be considered as a special set consisting of $\{A_1, A_2, ...\}$. According to Kolmogorov's zero-one law (\cite{fritz2020zero}), any element in $\mathcal A$ possesses the zero-one property. Since $\liminf_{i\to\infty}A_i\notin \mathcal A$, the zero-one law does not necessarily apply to the limit inferior; it becomes different on the leaf interval of causal trees. 
Combining \textbf{Theorem \ref{LILI-01}} and \textbf{Theorem \ref{LSLI-01}}, we establish that the zero-one law holds for both LSLI and LILI, even in the absence of belonging to the tail $\sigma$-field $\mathcal A$. Furthermore, we provide specific conditions under which the probability is either zero or one. 

In conclusion, LSLI and LILI satisfy nearly all the properties of the original limit superior and limit inferior, while transcending the original framework of the zero-one law on $\mathcal A$ and demonstrating greater symmetry through the unique characteristics of causal trees. However, due to its broad scope and tendency to amplify bias, LSLI is not suitable for ATE computation. We mainly focus on how to construct a set of LILIs with $K$ trees, $\mathcal I=\{I(\mathbf x_1, \Theta_K), I(\mathbf x_2, \Theta_K),...\}$, and derive the ATE.

To establish LILI as a valid clustering method, its well-definedness must be ensured. If $\mathbf y\in I(\mathbf x, \Theta_K)$, then $\mathbf y\in I(\mathbf x,\theta_i)$ on at least $K-[\sqrt{K}]$ trees. It also means $\mathbf x\in I(\mathbf y, \theta_i)$ on the same $K-[\sqrt{K}]$, which leads to $\mathbf x\in I(\mathbf y, \Theta_K)$. This verifies that $\mathbf y\in I(\mathbf x, \Theta_K)\Longleftrightarrow \mathbf x\in I(\mathbf y, \Theta_K)$, and it still holds when $K,n\to\infty$.
Moreover, all instances are uniquely classified into one LILI, ensuring that there is no overlap between different LILIs.

\begin{theorem}
\label{well-define}
Let $\mathbf x, \mathbf y$ be two different instances. If $P(\mathbf y\in I(\mathbf x))<1$, then $P(I(\mathbf x)\bigcap I(\mathbf y) = \emptyset)=1$. Moreover, $\mathbf x, \mathbf y$ are in the same LILI if and only if the LILIs of $\mathbf x,\mathbf y$ are equal in probability, i.e.
\begin{align*}
    P(\mathbf y\in I(\mathbf x)) =1 \Longleftrightarrow P(I(\mathbf x)=I(\mathbf y))=1.
\end{align*}
\end{theorem}

In practical applications, only a finite number of causal trees and intervals $I(\mathbf x, \Theta_K)$ can be constructed. As a result, some instances may belong to multiple $I(\mathbf x, \Theta_K)$. To address this issue, it is essential to clarify the exact form of the intersections of LILIs. For ordinary limit inferior of $A_i$ and $B_i$, $\liminf A_i\bigcap\liminf B_i=\liminf(A_i\bigcap B_i)$. Unfortunately, this intersection property of traditional limit inferior sets does not fully extend to LILI
\begin{align}
\label{LILI-intersection}
\bigcup_{1\leq i_1<\cdots i_s\leq K\atop s> K-\sqrt{K}}\bigcap_{k=1}^{s}(I(\mathbf x,\theta_{i_k})\cap I(\mathbf y,\theta_{i_k}))=I(\mathbf x, \mathbf y, \Theta_K)\subset I(\mathbf x,\Theta_K)\cap I(\mathbf y,\Theta_K).
\end{align}
In fact, if $\mathbf w,\mathbf x$ are in the same leaves on causal trees with indices $1,2, ..., K-[\sqrt{K}]$, and $\mathbf w,\mathbf y$ are in the same leaves on causal trees with indices $[\sqrt{K}]+1,..., K$, then $\mathbf x, \mathbf y$ are grouped into the same leaves at least on causal trees with indices $[\sqrt{K}]+1,..., K-[\sqrt{K}]$. Note that $\mathbf w\in I(\mathbf x,\mathbf y,\Theta_K)$ if only if $\mathbf w\in L(\mathbf x)\bigcap L(\mathbf y)$ on at least $K-[\sqrt{K}]$ trees, according to the same definition on $I(\mathbf x, \Theta_K)$. But on each tree, $\mathbf w$ can only belong to one leaf, then $\mathbf w\in L(\mathbf x)\bigcap L(\mathbf y)\Longleftrightarrow \mathbf w\in L(\mathbf x), L(\mathbf x)=L(\mathbf y)$. If $\mathbf x, \mathbf y$ are in the same leaves only on these $K-2[\sqrt{K}]$ trees, we just complete an example that $\mathbf w\in I(\mathbf x,\Theta_K)\bigcap I(\mathbf y,\Theta_K)$ but $\mathbf w\notin I(\mathbf x,\mathbf y,\Theta_K)$. Meanwhile, the discussion above also proves 
\begin{align*}
    I(\mathbf x,\Theta_K)\cap I(\mathbf y,\Theta_K)\subseteq \bigcup_{1\leq i_1<\cdots i_s\leq K\atop s> K-2\sqrt{K}}\bigcap_{k=1}^{s}(I(\mathbf x,\theta_{i_k})\cap I(\mathbf y,\theta_{i_k})).
\end{align*}
Under this relationship, we can investigate the convergence speed about the intersection of $I(\mathbf x, \Theta_K), I(\mathbf y,\Theta_K)$.
\begin{theorem}
\label{inter-convergence}
Given the finite data set $\mathcal D$, causal forest with $K$ trees $\Theta_K$ and two instances $\mathbf x,\mathbf y$, suppose $P(\mathbf y\in L(\mathbf x))=p<1$, then for any possible instance $\mathbf w\in\mathcal X$,
\begin{align}
\label{inter-bound}
    P(\mathbf w\in I(\mathbf x,\Theta_K)\cap I(\mathbf y,\Theta_K))\leq (pp_\mathbf w)^{K}+2(1-pp_\mathbf w)(pp_\mathbf w)^{K-2\sqrt{K}}\sqrt{K}K^{2\sqrt{K}},
\end{align}
where $p_\mathbf w = \min(P(\mathbf w\in L(\mathbf x)), P(\mathbf w\in L(\mathbf y)))$. Furthermore, 
\begin{align}
\label{convergence-speed}
\lim_{K\to\infty}\lim_{n\to\infty}\frac{P(\mathbf w\in I(\mathbf x,\Theta_K)\cap I(\mathbf y,\Theta_K))}{\min(p,p_\mathbf w)^{K-2\sqrt{K}}}= 0.
\end{align}
\end{theorem}

Equation (\ref{convergence-speed}) clarifies that $P(\mathbf w\ in I(\mathbf x, \Theta_K)\cap I(\mathbf y, \Theta_K))= o(\min(p,p_\mathbf w)^{K-2\sqrt{K}}$. Since $\mathbf w$ is an arbitrary instance, $P(I(\mathbf x, \Theta_K)\cap I(\mathbf y,\Theta_K)=\emptyset)=o(p^{K-2\sqrt{K}})$. Understanding this convergence speed is essential for assessing the reliability of the clustering method and its effectiveness in reducing bias in causal inference. 

Although any two LILIs with $K$ trees may have overlapping instances, this probability decreases exponentially as the number of causal trees increases. In practice, we can manually classify instances into 
$\mathcal I$ without replacement. This approach ensures that the classification remains unique. Thus, the ITE and the ATE for each $I(\mathbf x, \Theta_K)$ can be computed as 
\begin{align}
\widehat{ITE}_i &= 
\left\{
\begin{aligned}
Y_i -  \frac{1}{|I_c(\mathbf x, \Theta_K)|}\sum_{j\in I_c(\mathbf x, \Theta_K)} Y_j, \ T_i = 1;   \\
\frac{1}{|I_t(\mathbf x, \Theta_K)|}\sum_{j\in I_t(\mathbf x, \Theta_K)} Y_j - Y_i, \ T_i = 0;
\end{aligned}
\right. \\
\label{ATE_x}
    \widehat{ATE}(I(\mathbf x, \Theta_K)) &= \frac{1}{\left|I_t(\mathbf x,\Theta_K)\right|}\sum_{i\in I_t(\mathbf x,\Theta_K)} Y_i - \frac{1}{\left|I_c(\mathbf x,\Theta_K)\right|}\sum_{i\in I_c(\mathbf x,\Theta_K)} Y_i, 
\end{align}
where $I_t(\mathbf x, \Theta_K)$ and $I_c(\mathbf x, \Theta_K)$ are the treatment group and control group in $I(\mathbf x,\Theta_K)$. $\widehat{ATE}(I(\mathbf x, \Theta_K))$ is an estimation of conditional ATE (CATE), 
\begin{align}
\label{CATE}
    CATE(I(\mathbf x, \Theta_K)) = E(Y\mid T=1, I(\mathbf x, \Theta_K))-E(Y\mid T=0, I(\mathbf x, \Theta_K)).
\end{align}
Since each instance is assigned to only one LILI, the estimated probability that data falls into $I(\mathbf x, \Theta_K)$ is given by $\widehat P(I(\mathbf x, \Theta_K))= \left|I(\mathbf x, \Theta_K)\right|/n$. 
To eliminate $\mathbf X\in I(\mathbf x, \Theta_K)$ term in CATE, we use weighted average of $ATE(I(\mathbf x, \Theta_K))$ as the prediction of true ATE:
\begin{align}
\label{ATE}
    \widehat{ATE} = \sum_{I(\mathbf x,\Theta_K)\in \mathcal I}\widehat{ATE}(I(\mathbf x,\Theta_K))\widehat P(I(\mathbf x, \Theta_K)).
\end{align}
Currently, we present a detailed algorithm for computing the ATE using LILI, as outlined in \textbf{Algorithm \ref{LILI-alg}}.

\begin{algorithm}[H]
\caption{LILI clustering}
\label{LILI-alg}
\KwIn{$\mathcal D$, well-trained random forest $\Theta_K$}
\KwOut{$ATE$}
Initialize $\mathcal I=\emptyset$\;
\For{every instance $\mathbf x$ in $\mathcal D$}{
 $I(\mathbf x,\Theta_K)=\{\mathbf x\}$\;
\For{every instance $\mathbf y\neq \mathbf x$ in $\mathcal D$}{
$count = 0$\;
\For{causal tree $\theta_i$ in $\Theta_K$}{
\If{$\mathbf x,\mathbf y$ are in the same leaf on $\theta_i$}{
count +=1\;}
}
\If{count $>$ $K-\sqrt{K}$}{
Add $\mathbf y$ into $I(\mathbf x, \Theta_K)$\;}
}
Remove all the instances in $I(\mathbf x, \theta_K)$ from $\mathcal D$\;
Add $I(\mathbf x, \Theta_K)$ into $\mathcal I$\;}
\For {$I(\mathbf x, \Theta_K)$ in $\mathcal I$}{
\If{$I(\mathbf x, \Theta_K)$ only contains treatment or control group}{
 Remove $I(\mathbf x, \Theta_K)$ from $\mathcal I$\;}}
$n'= \sum_{I(\mathbf x, \Theta_K)\in \mathcal I}|I(\mathbf x, \Theta_K)|$\;
\For{$I(\mathbf x, \Theta_K)$ in $\mathcal I$}{
Compute $ATE(I(\mathbf x, \Theta_K))$ using equation (\ref{ATE_x})\;
$P(I(\mathbf x, \Theta))=|I(\mathbf x,\Theta_K)|/n'$\;}
Compute $ATE$ using (\ref{ATE});
\end{algorithm}

From line 3 to line 14, the LILI clustering algorithm generates a cluster around $\mathbf x$. This procedure traverses other instances $\mathbf y\in \mathcal D$ and evaluates whether $\mathbf y\in I(\mathbf x, \Theta_K)$ according to \textbf{Definition \ref{LILI}}. To avoid overlap between different $I(\mathbf x, \Theta_K)$, classified instances are removed from the dataset in line 15. After line 17, each instance is grouped into only one cluster $I(\mathbf x, \Theta_K)$, and all clusters are gathered into the set  $\mathcal I$. However, to compute $ATE(I(\mathbf  x, \Theta_K))$, it is essential that $I(\mathbf x, \Theta_K)$ contains both the treatment group and the control group. Lines 18-22 are designed to exclude the cluster that contains only one side. This implies that the LILI clustering algorithm identifies the samples in excluded clusters as outliers, and there are no counterfactual instances matching with them. This exclusion also helps mitigate bias, as not all instances are suitable for estimating ATE. Meanwhile, the total number of samples is recalculated as the sum of the samples in the remaining $I(\mathbf x, \Theta_K)$. Finally, the computation of $ATE$ follows equations \ref{ATE_x} and \ref{ATE}.

The computational complexity primarily arises from the construction of $I(\mathbf x,\Theta_K)$, with the complexity of traversing these tree layers being $O(n^2K)$. In the next section, we discuss the convergence of the estimated ATE from \textbf{Algorithm \ref{LILI-alg}} to the true ATE.

\section{Theoretical Analysis}
This section presents a theoretical framework demonstrating that the ATE derived from \textbf{Algorithm \ref{LILI-alg}} provides an unbiased estimation of the true ATE. We start by defining the estimated distribution induced by LILI using $K$ trees. After establishing the necessary assumptions and lemmas, we prove the convergence of the estimated distribution to the true distribution in probability. Next, we show that the estimated CATE in equation (\ref{CATE}) can be derived from the conditional expectation of a variable that follows the estimated distribution. Consequently, we establish the convergence of both the estimated CATE and estimated ATE to the ground truth. Additionally, this section extends the function $\sqrt{K}$ in the definition of LILI to encompass more general $f(K)$. Through theoretical analysis, we provide suggestions for fine-tuning $f(K)$.

As the LILI clustering uses the framework of random forest, it can also be treated as a prediction model for label $Y$. Let's define the weight of $i$th instances given $I(\mathbf x, \Theta_K)$ as 
\begin{align*}
    W_i(I(\mathbf x, \Theta_K)) = \frac{\mathbf{1}(\mathbf X_i\in I(\mathbf x,\Theta_K))}{|I(\mathbf x, \Theta_K)|},
\end{align*} 
where $\mathbf{1}$ is indicator function and $|I(\mathbf x, \Theta_K)|$ is number of instances belong to $I(\mathbf x, \Theta_K)$. Then, according to the estimation on ATE, the distribution of $Y$ induced by LILI, $F(y\mid I(\mathbf x, \Theta_K)) = P(Y\leq y\mid \mathbf X\in I(\mathbf x,\Theta_K))$, can be estimated as 
\begin{align}
\label{estimated-distribution}
    \widehat F(y\mid I(\mathbf x, \Theta_K))= \sum_{i=1}^n W_i(I(\mathbf x, \Theta_K))\mathbf{1}(Y_i\leq y),
\end{align}

It is trivial that $\sum^n_{i=1}W_i(I(\mathbf x, \Theta_K))=1$. Similar to other distributions in causal forest methods, such as those discussed in \cite{meinshausen2006quantile, cevid2022distributional}, some assumptions are required for the convergence of the estimated distribution.

\begin{itemize}
    \item \textit{Lipschitz continuity}: For the true distribution of $Y$, $F(Y\mid \mathbf X=\mathbf x)$ and any two instances $\mathbf x$, $\mathbf x'$, there is a constant $L$ such that
    \begin{align*}
        \sup_y|F(y\mid\mathbf X=\mathbf x)-F(y\mid\mathbf X=\mathbf x')|\leq L\parallel\mathbf x-\mathbf x'\parallel_1,
    \end{align*}
    where $\parallel\cdot\parallel_1$ is L1 norm;
    \item \textit{monotonicity}: The distribution $F(\mathbf y\mid \mathbf X=\mathbf x)$ is strictly monotonously increasing with respect to $y$ for any given $\mathbf x$.
\end{itemize}
The convergence properties outlined in \cite{meinshausen2006quantile, cevid2022distributional} typically require more stringent assumptions than those we consider here. We focus on two common assumptions and exclude the rest. Given the strong properties of $I(\mathbf x, \Theta_K)$ established in \textbf{Section 3}, it is sufficient to verify the convergence of estimation (\ref{estimated-distribution}) under \textit{Lipschitez continuous} and \textit{monotonicity}. 
Firstly, we present the necessary lemmas.
\begin{lemma}
\label{LILI-measure}
Denote $\mu$ as the Borel measure on the real number field and $\mathcal B$ as the range of instances. Assume that $\mu(\mathcal B)<+\infty$, and for any continuous variables, there are no $l$ instances that have the same value. Then $\eta$, the superior of decrease proportion on adding a split node, has
\begin{align*}
    \eta \triangleq \sup_{\mathbf x, \theta_k, c}\min_j\frac{\mu(I(\mathbf x, \theta_k, c)_j)}{\mu(I(\mathbf x, \theta_k, c+1)_j)} < 1,
\end{align*}
where $I(\mathbf x, \theta_k, c)$ is the interval on the $c$th level of the $k$th tree containing $\mathbf x$, and $I(\mathbf x, \theta_k, c)_j$ is its $j$th dimension. And
\begin{align}
\label{l1-convergence-speed}
    P\left(\parallel \mathbf y-\mathbf w\parallel_1 >\epsilon\mid \mathbf y, \mathbf w\in I(\mathbf x, \Theta_K)\right)\leq \frac{1}{\epsilon^2}\eta^{\pi K (\ln n-\ln l)/(-d\ln\alpha)}\mu(\mathcal B).
\end{align}
\end{lemma}

Borel measure is used on the real field $\mu([a,b]) = b-a$. Recall in the beginning of \textbf{Section 3}, we analyze that the leaf interval of the $k$th tree has the form $I(\mathbf x, \theta_k)=\prod_{i=1}^d(a_i^k, b_i^k)$. As we know, the union or intersection of two different intervals $(\tilde a,\tilde b), (a',b')$ can be one interval $(a,b)$ as long as they have an overlapping part. Although $I(\mathbf x, \Theta_K)$ is the union and intersection of different leaf intervals, they all contain $\mathbf x$, and any two of them have overlapping parts. Hence, $I(\mathbf x, \Theta_K)$ can be represented as $\prod_{i=1}^d(a_i, b_i)$, and 
\begin{align*}
    \mu(I(\mathbf x,\Theta_K)) = \sum^d_{i=1} (b_i-a_i) = \sup_{\mathbf y, \mathbf w\in I(\mathbf x, \Theta_K)}\parallel\mathbf y-\mathbf w\parallel_1
\end{align*}
which makes $\mu$ a perfect measure for the distance of $I(\mathbf x,\Theta_K)$. Consequently, the boundary of inequality (\ref{l1-convergence-speed}) is also the boundary of probability on $\mu(I(\mathbf x, \Theta_K))$. 

\begin{lemma}
\label{LILI-size}
Let $q_\mathbf x$ be the probability that samples belong to leaf $L(\mathbf x)$ under infinity samples, 
$q_\mathbf x \triangleq P(L(\mathbf x)) = \lim_{n\to\infty}\sum_{\mathbf y}P(\mathbf y\in L(\mathbf x))P(\mathbf y)$, then $\alpha /2< q_\mathbf x <2\alpha$.
If the number of instances $n=O\left((2/\alpha)^K\right)$,i.e. $n(\alpha/2)^K\to\infty$, we have 
\begin{align*}
    |I(\mathbf x, \Theta_K)|\overset{P}{\to} +\infty.
\end{align*}
\end{lemma}

In real-world datasets, it is uncommon for two instances to share identical values for continuous variables. Taking into account the \textit{$\alpha$-regularity}, \textbf{Lemma \ref{LILI-measure}} generalizes to $l$ instances within its assumption. As we consider the limits of both $n$ and $K$, varying relationships between $n$ and $K$ can yield different limiting values. \textbf{Lemma \ref{LILI-size}} establishes that the size of $I(\mathbf x, \Theta_K)$ can approach infinity if the dataset's growth rate is exponential with respect to $K$, which can be controlled by adjusting $K$ in experiments. Therefore, the assumptions in \textbf{Lemma \ref{LILI-measure}} and \textbf{Lemma \ref{LILI-size}} are feasible for real-world applications. With these two lemmas, we can demonstrate the convergence of estimation (\ref{estimated-distribution})

\begin{theorem}
\label{estimation-convergence}
Under the condition of \textbf{Lemma \ref{LILI-measure}} and \textbf{Lemma \ref{LILI-size}}, $\widehat F(y\mid I(\mathbf x, \Theta_K))$ converges to $F(y\mid I(\mathbf x, \Theta_K))$ in probability. Furthermore, 
\begin{align}
\label{estimation-convergence-speed}
    P(|\widehat F(y\mid I(\mathbf x, \Theta_K)) - F(y\mid I(\mathbf x, \Theta_K))| \geq \epsilon) \leq \frac{1}{\epsilon^2}\left(nq_\mathbf x^K + L\eta^{\pi K (\ln n-\ln l)/(-d\ln\alpha)}\mu(\mathcal B)\right).
\end{align}
\end{theorem}

Besides these, we can also analyze the size of $\mathcal I$ according to \textbf{Lemma \ref{LILI-size}}.  In fact, regardless of whether the assumption from \textbf{Lemma \ref{LILI-size}} is applied, the relationship $|I(\mathbf x, \Theta_K)|\geq nq_\mathbf x^K$ holds, and 
\begin{align*}
    n = \sum_{I(\mathbf x, \Theta_K)\in \mathcal I} |I(\mathbf x, \Theta_K)|\geq |\mathcal I|\min |I(\mathbf x, \Theta_K)| \geq |\mathcal I| nq_\mathbf y^K,
\end{align*}
where $\mathbf y$ is the instance that $|I(\mathbf y, \Theta_K)|$ reaches its minimum in $\mathcal I$. Hence, $|\mathcal I|\leq \left(\frac{1}{q_\mathbf y}\right)^K$. From the definition of LILI, the number of clusters is solely influenced by the number of causal trees. The analysis above states that the growth rate of $|\mathcal I|$ is restricted by an exponential function of $K$.


Recall that $I(\mathbf x, \Theta_K)$ is a cluster centered around the covariates $\mathbf X$, and $\widehat F(y\mid I(\mathbf x, \Theta_K))$ denotes the conditional distribution within this cluster. To clarify the convergence of the estimated ATE (\ref{ATE_x}), we extend the definition of equation (\ref{estimated-distribution}) to encompass both the treatment and control groups, 

\begin{align}
    \widehat F(y\mid T=1, I(\mathbf x, \Theta_K)) \triangleq \sum^n_{i=1} \frac{\mathbf 1(\mathbf X\in I_t(\mathbf x, \Theta_K))}{|I_t(\mathbf x, \Theta_K)|}\mathbf 1(Y_i\leq y).
\end{align}
For $\widehat F(y\mid T=0, I(\mathbf x, \Theta_K))$, $I_t(\mathbf x, \Theta_K)$ is replaced by $I_c(\mathbf x, \Theta_K)$. As $I_t(\mathbf x, \Theta_K), I_c(\mathbf x, \Theta_K) \subset I(\mathbf x, \Theta_K)$, 
\begin{align*}
    &P\left(\parallel \mathbf y-\mathbf w\parallel_1 >\epsilon\mid \mathbf y, \mathbf w\in I_t(\mathbf x, \Theta_K)\right) \leq
    P\left(\parallel \mathbf y-\mathbf w\parallel_1 >\epsilon\mid \mathbf y, \mathbf w\in I(\mathbf x, \Theta_K)\right).
\end{align*}
Furthermore, by the \textit{Positivity} assumption in \textbf{Section 2}, $P(T\mid \mathbf X\in I(\mathbf x, \Theta_K))>0$, then with the growing number of instances, 
\begin{align*}
    |I_t(\mathbf x, \Theta_K)|\to P(T=1\mid \mathbf X\in I(\mathbf x, \Theta_K))|I(\mathbf x, \Theta_K)|.
\end{align*}
$I_c(\mathbf x, \Theta_K)$ also has the same properties as described in the two equations above. Following the same proof procedure, both \textbf{Lemma \ref{LILI-measure}} and \textbf{Lemma \ref{LILI-size}} hold for $I_t(\mathbf x, \Theta_K)$ and $I_c(\mathbf x, \Theta_K)$. Therefore, the convergence of these two estimations can be established using the approach outlined in \textbf{Theorem \ref{estimation-convergence}},  thereby ensuring the convergence of $\widehat{ATE}$.

\begin{theorem}
\label{ATE-convergence}
With the condition in \textbf{Theorem \ref{estimation-convergence}}, it holds that
\begin{align*}
    \widehat{ATE}(I(\mathbf x, \Theta_K)) &\overset{P}{\to} CATE(I(\mathbf x, \Theta_K)). \\
    \widehat{ATE} &\overset{P}{\to} ATE
\end{align*}
with respect to $n,K$.
\end{theorem}

All previous discussions are based on the definition of LILI in \textbf{Definition \ref{LILI}}. The function $\sqrt{K}$ serves as a tolerance, representing the degree to which LILI representing the degree to which LILI allows samples to reside in different leaves. 
For convenience, we refer to $\sqrt{K}$ as the tolerance function. This raises the question: 
what happens if the tolerance function is replaced with alternative functions of $K$? 
Hence, it is essential to explore whether LILI retains similar properties and whether the estimation in (\ref{estimated-distribution}) converges with different tolerance functions.

First of all, we extend the definition of LILI to accommodate different tolerance functions. The LILI with $K$ trees under the tolerance function $f$ is defined as
\begin{align}
    I(\mathbf x, \Theta_K, f) = \bigcup_{1\leq i_1<\cdots i_s\leq K\atop s> K-f(K)}\bigcap_{k=1}^{s} I(\mathbf x, \theta_{i_k}).
\end{align}
Similarly, the LILI under $f$ is shown as $I(\mathbf x, f) = \lim_{K\to\infty}\lim_{n\to\infty} I(\mathbf x, \Theta_K, f)$. Apparently, $f(K)\leq K$, and when $f(K)=\sqrt{K}$, $I(\mathbf x, f) = I(\mathbf x)$.
The two extreme situations need to be excluded. For $f(K) = K$ and $f(K) = 0$, 
\begin{align*}
    I(\mathbf x, \Theta_K, K) = \bigcap_{k=1}^K I(\mathbf x, \theta_k),\ \ \ \  I(\mathbf x, \Theta_K, 0) = \bigcup_{k=1}^K I(\mathbf x, \theta_k),
\end{align*}
they have different formations from limit inferior. In practice, $I(\mathbf x, \Theta_K, K)$ typically contains no sample other than $\mathbf x$ itself, whereas $I(\mathbf x, \Theta_K, 0)$ includes all samples. Therefore, the following discussion restricts $f$ to satisfy $0<f(K)<K$.

\begin{theorem}
\label{LILI-extention1}
For given $\mathbf x, \mathbf y$, let $p = \lim_{n\to\infty}P(\mathbf y\in L(\mathbf x))<1$, then $ P(\mathbf y\in I(\mathbf x,f)) = 0$ achieves if 
\begin{align}
\label{f-require1}
    \lim_{K\to\infty}\frac{f(K)\ln K}{K} < -\ln p.
\end{align}
Furthermore, if $p<0.5$, $P(\mathbf y\in I(\mathbf x,f))=0$ only requires \begin{align}
\label{f-require4}
    \lim_{n\to\infty}\frac{f(K)}{K} = 0.
\end{align}
And for any $f$, $\lim_{n\to\infty}P(\mathbf y\in L(\mathbf x)) = 1 \Longrightarrow P(\mathbf y\in I(\mathbf x, \Theta_K, f)) = 1$.
\end{theorem}

To achieve the zero-one law in \textbf{Corollary \ref{LILI-01}}, there are multiple choices for $f(K)$, as long as it satisfies (\ref{f-require1}). However, if the probability $\lim_{n\to\infty}P(\mathbf y\in L(\mathbf x)) < 0.5$, this restriction can be expanded such that $f(K)/K\to\infty$. Conversely, if we aim to select $\mathbf y$ such that $\mathbf y\in P(\mathbf y\in L(\mathbf x))>0.5$, the tolerance function only needs to meet (\ref{f-require4}). Then for $f(K)$ in (\ref{f-require1}), the equivalence (\ref{LI-equivalence}) holds, i.e.
\begin{align*}
    P(y\in I(\mathbf x,f))=1 \Longleftrightarrow \lim_{n\to\infty}P(\mathbf y\in L(\mathbf x))=1
\end{align*}
as well as the non-overlapping property in \textbf{Theorem \ref{well-define}}. 
In section 3, we aim to identify the counterfactual instances $\mathbf y$ such that $\lim_{n\to\infty}P(\mathbf y\in L(\mathbf x)) = 1$. Conversely, if we want to expand this range, the tolerance function can serve as a controlling factor. By amplifying $f$, it is possible to include more instances with $\lim_{n\to\infty}P(\mathbf y\in L(\mathbf x)) \leq 1$ into $I(\mathbf x, f)$. However, this amplification is constrained and has its limits.

\begin{theorem}
\label{LILI-extention2}
For tolerance function $f$ that satisfies 
\begin{align}
\label{f-require2}
    \lim_{n\to\infty} \frac{K-f(K)}{\ln K} = 0,
\end{align}
then $p = P(\mathbf y\in L(\mathbf x)) >0\Longrightarrow P(\mathbf y\in I(\mathbf x, f))=1$. Furthermore, if $p > 0.5$, the tolerance function only needs to satisfy 
\begin{align}
\label{f-require3}
    \lim_{K\to\infty}\frac{K}{(K-f(K))2^{K-f(K)}} = +\infty,
\end{align}
to achieve $P(\mathbf y \in I(\mathbf x, f))=1$
\end{theorem}

\textbf{Theorem \ref{LILI-extention1}} and \textbf{Theorem \ref{LILI-extention2}} provide guidance on adjusting the tolerance function. To preserve all the properties of the original LILI, the tolerance function must remain within the range specified by (\ref{f-require1}). If we need to disregard the zero-one law and include more samples in the cluster $I(\mathbf x, K, f)$, $f$ cannot fall within the range specified by (\ref{f-require2}). In fact, $I(\mathbf x, K, f)$ with the tolerance function in (\ref{f-require2}) follows the zero-one law in \textbf{Theorem \ref{LSLI-01}} and the equivalence in (\ref{LS-equivalence}). This implies that when $f$ becomes sufficiently large, $I(\mathbf x, f)$ has the same properties as $\bar I(\mathbf x)$. In such cases, $I(\mathbf x,f)$ may include all samples for which $P(\mathbf y\in L(\mathbf x)) > 0$, leading to significant errors in the estimated ATE.

\begin{figure*}
\centering
\includegraphics[scale=0.7]{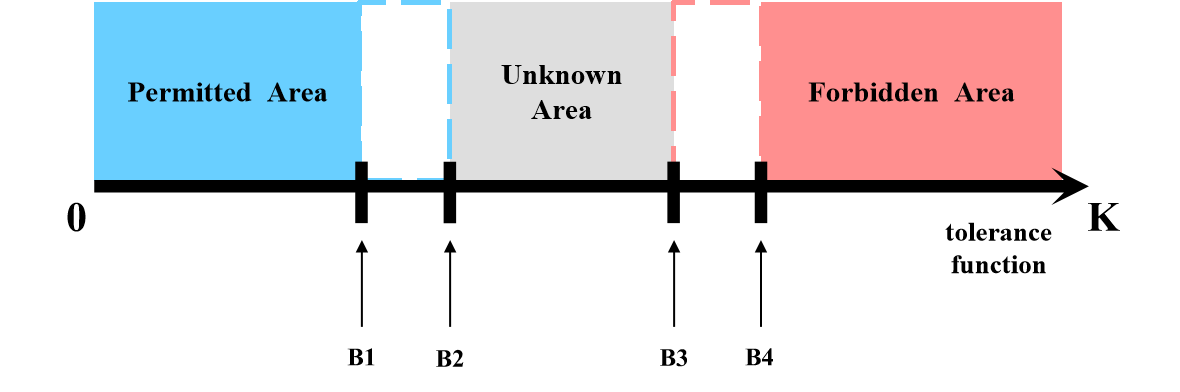}
\caption{The areas for adjusting the tolerance function. Four boundaries B1, B2, B3, B4 are defined by equations (\ref{f-require1}), (\ref{f-require4}), (\ref{f-require2}), and (\ref{f-require3}). The $I(\mathbf x, f)$ under the tolerance function within the permitted area satisfies all the properties of $I(\mathbf x)$. While any $f$ in the forbidden area fails to uphold these properties. If the objective is to identify $\mathbf y$ with $\lim_{n\to\infty}P(\mathbf y\in L(\mathbf x))>0.5$, the permitted and forbidden area can be extended to B2 and B3 (the white area with dashed borders). The behavior of the tolerance function within the interval [B2, B3] remains unknown with current theorems.}\label{f-area}
\end{figure*}


To ensure probability convergence on the ATE,
the estimated distribution $F(y\mid I(\mathbf x, f))$ induced by $I(\mathbf x, \Theta_K, f)$ must also achieve probability convergence as outlined in \textbf{Theorem \ref{estimation-convergence}}. According to the proof, $I(\mathbf x, f)$ only needs to satisfy the  two lemmas required in \textbf{Theorem \ref{estimation-convergence}}. 

A larger tolerance function increases the measure of $I(\mathbf x,K, f)$, allowing it to encompass more samples.
For the tolerance function in (\ref{f-require2}), it is obvious that \textbf{Lemma \ref{LILI-size}} holds for $I(\mathbf x,\Theta_K,f)$ as $|I(\mathbf x, \Theta_K, f)|\geq |I(\mathbf x, \Theta_K, f)|$. Referring to the analysis in equation (\ref{LILI-intersection}), for two samples $\mathbf y, \mathbf w\in I(\mathbf x,\Theta_K)$, $\mathbf x, \mathbf y, \mathbf w$ belong to the same leaves for at least $K-2[\sqrt{K}]$, given that $K-[\sqrt{K}]\gg K/2$ for sufficiently large $K$.
However, for $K-f(K)\ll K/2$ with $f$ in (\ref{f-require2}), it is possible for two samples $\mathbf w, \mathbf y$ to fall into entirely different intersections of leaf nodes. For example,
\begin{align*}
    \mathbf y\in \bigcap_{k=1}^{K-[f(K)]} I(\mathbf x,\theta_k), \ \ \ \ \ \ 
    \mathbf w\in \bigcap_{k=K-[f(K)]+1}^{2K-2[f(K)]} I(\mathbf x, \theta_k), \ \ \ \ \ \ 
    \mathbf y, \mathbf w\in I(\mathbf x, \Theta_K, f).
\end{align*}
Therefore, it becomes challenging to constrain the distance between two samples, $\parallel \mathbf y-\mathbf w\parallel_1$, as well as the $\mu(I(\mathbf x, \Theta_K, f))$. Overall, $I(\mathbf x, \Theta_K, f)$ with $f$ in (\ref{f-require2}) to satisfy \textbf{Lemma \ref{LILI-measure}}, causing the estimated distribution induced by it to diverge from the true distribution.

On the other hand, when $f$ satisfies (\ref{f-require1}), $f(K)\ll K/2$, $I(\mathbf x, \Theta_K, f)$ adheres to the \textbf{Lemma \ref{LILI-measure}}. According to the \textit{$\alpha$-regularity}, the leaf nodes contain at least $\alpha n$ samples. Therefore, regardless of how small the tolerance function is, \textbf{Lemma \ref{LILI-size}} is upheld as long as \textit{$\alpha$-regularity} exists. In summary, when the tolerance function lies within the range specified in (\ref{f-require1}), all the relevant properties and theorems are achieved. On the contrary, when $f$ in (\ref{f-require2}), $I(\mathbf x, \Theta_K, f)$ fails to meet most critical properties, 
marking (\ref{f-require2}) as a forbidden area for tuning $f$. The visualization of these areas is presented in Figure \ref{f-area}.

\section{Experiment}

In this section, we first outline the experimental setup, including the experimental setup, including details on the datasets, data preprocessing procedures, and the hyperparameters used in both the baseline algorithms and the LILI clustering method. Next, we evaluate the proposed algorithm on these datasets in comparison to the baseline methods, with a focus on estimating both the ATE and the ITE. 
We further analyze the impact of key hyperparameters, such as the number of trees $K$, and $\alpha, l$ in \textit{$\alpha$-regularity}. Lastly, we propose a detailed approach for fine-tuning the LILI clustering process.

\subsection{Experimental Setup}

In the study of causal inference,  one of the primary challenges is the absence of ground truth for counterfactual outcomes and treatment effects. Therefore, most datasets used are semi-synthetic. For the following experiments, we utilize semi-synthetic datasets, including the Infant Health and Development Program (IHDP) dataset (\cite{brooks1992effects}), the Atlantic Causal Inference Conference (ACIC) challenge datasets from 2016 (ACIC16) (\cite{ACIC16}) and 2019 (ACIC19) (\cite{ACIC19}), as well as the real-world TWINS dataset (\cite{almond2005costs}). Table \ref{data-info} provides the basic information and hyperparameter settings for these datasets.

\begin{table*}[htbp]
\centering
\fontsize{8}{14}\selectfont
\caption{Basic information and hyperparameters about datasets}\label{data-info}
\setlength{\tabcolsep}{3mm}{
\begin{tabular}{|c|c|c|c|c|c|}
\hline
\multirow{2}{*}{Dataset} & Dataset   & Dimension of   & Minimum & Tree   & Minimum  \cr  
 & Size   & Instance      &  Split proportion          & Size   & Leaf Instances    
\cr\hline 
\hline
 IHDP (1-5)    & 1492    & 25     & 0.1    & 50    & 149      \cr\hline
 ACIC16 (1-5)  & 4805    & 58     & 0.02   & 50    & 100      \cr\hline
 ACIC19 (1-6)  & 500     & 15     & 0.1    & 70    & 50       \cr\hline
 TWINS         & 2178    & 50     & 0.02   & 100   & 50       \cr\hline

\end{tabular}}
\end{table*}
The columns describe key details about each sub-dataset in IHDP, ACIC, and TWINS, including minimum split instances and minimum leaf instances, which correspond to the hyperparameters $\alpha, l$ in $\textit{$\alpha$-regularity}$, and tree size, which refers to $K$ in $I(\mathbf x, \Theta_K)$.

\textbf{Data description}: The IHDP is one of the most commonly used dataset for causal inference. It is generated by the randomized controlled experiment conducted in \cite{brooks1992effects}. The dataset consists of 10 sub-datasets, and in this study, every two sub-datasets are combined to form a new dataset. As a result, IHDP (1-5) contains all instances from the original 10 sub-datasets.
The ACIC16 and ACIC19 datasets, sourced from the Atlantic Causal Inference Conference, consist of numerous sub-datasets with covariates based on real-world data, while their factual and counterfactual outcomes are generated via simulation. For this experiment, the first five sub-datasets (ACIC16-1 to ACIC16-5, ACIC19-1 to ACIC19-5) are selected.
The TWINS dataset is a fully real-world dataset designed to investigate the effect of infant birth weight on infant mortality. It contains data on twins born in the USA between 1989 and 1991. Since twins share identical covariates, the lighter and heavier twins in each pair serve as counterfactual outcomes for one another, enabling a ground truth ATE calculation similar to that used with the IHDP dataset.

\textbf{Data Processing}: Discrete and continuous covariates are preprocessed using max-min normalization and label encoding. In the IHDP and ACIC datasets, the treatment variables are binary, whereas the weight variables in the TWINS dataset are continuous.

To investigate the causal relationship between weight and mortality, we first define a weight gap. Pairs of twins with a weight difference smaller than this threshold are excluded, as twins with similar weights do not provide meaningful causal information. Subsequently, we randomly select one infant from each twin pair, assigning $T=1$ if the selected infant is heavier and $T=0$ otherwise. The unselected infant serves as the counterfactual outcome for the selected one.
The IHDP datasets also provide counterfactual outcomes for each instance, enabling the calculation of ground truth ATEs. Using these counterfactual outcomes, we compute the ground truth ATEs for both the IHDP and TWINS datasets as follows:
\begin{align*}
    ATE = \frac{|\mathcal D_t|}{|\mathcal D|}\sum_{i\in\mathcal D_t}\left(Y_i - Y_i^{C}\right) + \frac{|\mathcal D_c|}{|\mathcal D|}\sum_{i\in \mathcal D_c} \left(Y_i^C-Y_i\right),
\end{align*}
where $\mathcal D_t$ ($\mathcal D_c$) are treatment (control) group in $\mathcal D$. While, ACIC directly provides all the true ATE for each sub-dataset.

\textbf{Baseline algorithms}: There are total 11 baseline algorithms in experiments. 
\begin{itemize}
    \item The meta-learning methods: Single-learner (S-learner), Two learner (T-learner) (\cite{kunzel2019metalearners}), Residual-learner (R-learner) (\cite{nie2021quasi}), and double machine learning (DML) (\cite{chernozhukov2018double}). The base learners in all meta-learning methods are set as random forests;
    \item Tree-based methods: Causal Forest (\cite{wager2018estimation}), Uplift Forest (\cite{rzepakowski2012decision}). The hyperparameters of tree-based methods mentioned in Table \ref{data-info} are the same as those used in LILI clustering;
    \item Re-weighting methods: Inverse propensity weighting (IPW) (\cite{rosenbaum1987model}), Doubly Robust Instrumental Variable (DRIV) learner, Targeted maximum likelihood estimation (TMLE) (\cite{van2011targeted}). DRIV-learner and TMLE use regression models to compute the weights for every instance. They can be viewed as combinations of re-weighting and meta-learning methods, but the frameworks of both methods still rely on the re-weighting of instances. 
    \item  Matching methods: Randomized Nearest Neighbor Matching (RNNM) (\cite{li2016matching}), Optimal Pair Matching (Optim) (\cite{hansen2006optimal}). For a specific instance in RNNM, we match five samples from different groups based on Euclidean distance.
\end{itemize}
\texttt{causalml} (\cite{zhao2023causal}) is a powerful causal inference package in Python that we use to implement S-learner, T-learner, R-learner, Causal Forest, Uplift Forest, IPW, DRIV-learner methods. DML is implemented by the Python package \texttt{doubleml} (\cite{bach2022doubleml}). Besides, we also apply R package \texttt{tmle} (\cite{gruber2012tmle}), \texttt{MatchIt} (\cite{ho2018package}) for TMLE, Optim, and RNNM algorithms. 

\textbf{Setup for LILI clustering}: The causal trees in LILI clustering use Euclidean distance as the split criterion, aiming to identify the optimal split features and split values by solving
\begin{align*}
    \min\sum_{i\in C_L} (Y_i - \bar Y_L)^2 + \sum_{i\in C_R}(Y_i-\bar Y_R)^2,
\end{align*}
where $C_L$ and $C_R$ are the left and right children nodes divided by the selected split feature and value on the current split node, and $\bar Y_L$, $\bar Y_R$ are the mean values of labels on $C_L$ and $C_R$. Similar to the original causal forest, LILI clustering also follows the four construction rules in \textbf{Section 2}. Except for $\alpha$, $l$ in \textit{$\alpha$-regularity} mentioned in Table \ref{data-info}, the hyperparameters in \textit{Data sampling} and \textit{Random-split} are fixed across all experimental datasets: $s_n = 80\%n$, $\pi = 0.1$. In the tables or graphs of the rest of this section, "LILI" is used as shorthand for "LILI clustering."

\subsection{Results On ATE}


\textbf{Algorithm \ref{LILI-alg}} is applied directly to the datasets under the experimental setup described above. For all baseline algorithms, we vary the random seed for each model and repeat the experiments five times for each dataset. The absolute error between the estimated and ground truth ATE, calculated as $|\widehat{ATE} - ATE|$, is used to compute the $L_1$ loss. We report the mean and standard deviation of $L_1$ loss as \textit{mean(standard deviation)} in Tables \ref{L1-IHDP}, \ref{L1-ACIA16}, and \ref{L1-ACIC19}.
It is worth noting that the matching algorithms, Optim and RNNM, are deterministic and do not involve randomness; therefore, their standard deviations are zero. For each sub-dataset, the lowest $L_1$ loss among all causal algorithms is highlighted in bold.

\begin{table*}[htpb]
\centering
\fontsize{10}{13}\selectfont
\caption{$L_1$ losses of ATE on IHDP datasets}
\label{L1-IHDP}
\setlength{\tabcolsep}{3mm}{
\begin{tabular}{cccccc}
\bottomrule[1.5pt]
\hline
\multirow{2}{*}{Algorithms} & \multirow{2}{*}{IHDP1} & \multirow{2}{*}{IHDP2} & \multirow{2}{*}{IHDP3} & \multirow{2}{*}{IHDP4} & \multirow{2}{*}{IHDP5}  \cr
 & \cr
\bottomrule
S-learner     & 0.250 (0.11) & 0.042 (0.02) & 0.025 (0.02) & 0.034 (0.01) & 1.122 (0.15) \cr
T-learner     & 0.072(0.09) & 0.017(0.01) & 0.021(0.01) & 0.041(0.02) & 0.243(0.07) \cr
R-learner     & 0.861(0.12) & 1.116(0.17) & 1.210(0.10) & 1.284(0.11) & 1.925(0.54) \cr
DML           & 0.133(0.06) & 0.047(0.03) & 0.053(0.04) & 0.049(0.02) & 0.108(0.02) \cr
Causal Forest & 0.267(0.10) & 0.066(0.01) & \textbf{0.020(0.01)} & 0.018(0.01) & 0.822(0.35) \cr
Uplift Forest & 1.064(0.04) & 1.763(0.01) & 5.416(0.01) & 0.081(0.01) & 3.162(0.09) \cr
IPW           & 0.149(0.00) & 0.058(0.00) & 0.028(0.00) & 0.021(0.00) & 0.654(0.00) \cr
DRIV-learner  & 0.645(0.41) & 0.241(0.11) & 0.171(0.08) & 0.265(0.08) & 2.547(1.59) \cr
TMLE          & 0.334(0.03) & 0.120(0.02)  & 0.103(0.02) & 0.041(0.01) & 2.604(0.13) \cr
Optim         & 0.086(0.00) & 0.023(0.00) & 0.090(0.00) & \textbf{0.017(0.00)} & 0.291(0.00) \cr
RNNM           & 0.128(0.00) & 0.105(0.00) & 0.277(0.00) & 0.052(0.00) & 0.899(0.00) \cr
LILI          & \textbf{0.048(0.03)} & \textbf{0.015(0.01)} & 0.076(0.01) & 0.049(0.01) & \textbf{0.015(0.01)} \cr
 \bottomrule[1.5pt]
\end{tabular}}
\end{table*}

\begin{table*}
\centering
\fontsize{10}{13}\selectfont
\caption{$L_1$ losses of ATE on ACIC16 datasets}
\label{L1-ACIA16}
\setlength{\tabcolsep}{3mm}{
\begin{tabular}{cccccc}
\bottomrule[1.5pt]
\hline
\multirow{2}{*}{Algorithms} & \multirow{2}{*}{ACIC16-1} & \multirow{2}{*}{ACIC16-2} & \multirow{2}{*}{ACIC16-3} & \multirow{2}{*}{ACIC16-4} & \multirow{2}{*}{ACIC16-5}  \cr
 & \cr
\bottomrule
S-learner     & 0.423(0.01) & 0.175(0.01) & 0.130(0.02) & 0.14(0.01) & 0.114(0.0)  \cr
T-learner     & 0.445(0.01) & 0.174(0.01) & 0.162(0.01) & 0.154(0.01) & 0.164(0.01) \cr
R-learner     & 0.791(0.03) & 0.175(0.01) & 0.279(0.04) & 0.222(0.02) & 0.136(0.01) \cr
DML           & 0.440(0.00) & 0.195(0.02) & 0.044(0.01) & 0.137(0.02) & 0.178(0.02)  \cr
Causal Forest & 1.517(0.02) & 1.201(0.01) & 1.936(0.05) & 0.624(0.02) & 1.306(0.02)  \cr
Uplift Forest & 3.102(0.01) & 2.161(0.01) & 3.63(0.02) & 3.513(0.01) & 0.753(0.02)  \cr
IPW           & 1.187(0.00) & 0.351(0.00) & 0.559(0.00) & 0.224(0.00) & 1.536(0.00)  \cr
DRIV-learner  & 0.190(0.05) & 0.324(0.03) & 0.096(0.03) & 0.10(0.01) & 0.252(0.03)  \cr
TMLE          & 0.222(0.01) & \textbf{0.014(0.01)} & 0.052(0.01) & 0.139(0.12) & 0.167(0.02) \cr
Optim         & 0.697(0.00) & 0.443(0.00) & 0.443(0.00) & 0.511(0.00) & 1.402(0.00)  \cr
RNNM           & 2.157(0.00) & 0.153(0.00) & 0.054(0.00) & 0.493(0.00) & 0.862(0.00)  \cr
LILI          & \textbf{0.113(0.01)} & 0.116(0.01) & \textbf{0.031(0.02)} & \textbf{0.031(0.03)} & \textbf{0.105(0.01)} \cr
 \bottomrule[1.5pt]
\end{tabular}}
\end{table*}

\begin{table*}
\centering
\fontsize{10}{13}\selectfont
\caption{$L_1$ losses of ATE on ACIC19 datasets}
\label{L1-ACIC19}
\setlength{\tabcolsep}{3mm}{
\begin{tabular}{ccccccc}
\bottomrule[1.5pt]
\hline
\multirow{2}{*}{Algorithms} & \multirow{2}{*}{ACIC19-1} & \multirow{2}{*}{ACIC19-2} & \multirow{2}{*}{ACIC19-3} & \multirow{2}{*}{ACIC19-4} & \multirow{2}{*}{ACIC19-5} & \multirow{2}{*}{ACIC19-6}  \cr
 & \cr
\bottomrule
S-learner     & 0.078(0.01) & 0.66(0.06) & 1.342(0.09) & 0.014(0.01) & 0.084(0.01) & 0.024(0.01) \cr
T-learner     & 0.017(0.01) & 0.294(0.02) & \textbf{0.253(0.05)} & 0.035(0.01) & 0.003(0.01) & 0.030(0.01)  \cr
R-learner     & 0.692(0.22) & 0.834(0.12) & 0.533(0.30) & 0.06(0.06) & 0.47(0.14) & 0.021(0.01)   \cr
DML           & 0.254(0.16) & 0.665(0.55) & 4.298(1.38) & 0.035(0.02) & 0.044(0.02) & 0.082(0.05)  \cr
Causal Forest & 0.081(0.01) & 2.376(0.10) & 7.077(0.22) & 0.016(0.01) & 0.050(0.01) & 0.027(0.01)  \cr
Uplift Forest & 0.119(0.01) & 6.866(0.05) & 9.447(0.07) & 0.055(0.01) & 0.180(0.01) & 0.065(0.01)  \cr
IPW           & 0.022(0.00) & 1.258(0.00) & 1.874(0.00) & 0.020(0.00) & 0.052(0.00) & 0.050(0.00)  \cr
DRIV-learner  & 0.019(0.01) & 0.360(0.03) & 0.481(0.03) & 0.038(0.01) & 0.016(0.01) & 0.037(0.01)  \cr
TMLE          & 0.018(0.01) & 1.311(0.05) & 3.228(0.66) & 0.041(0.01) & 0.098(0.02) & 0.041(0.01) \cr
Optim         & 0.051(0.00) & 0.125(0.00) & 0.787(0.00) & 0.042(0.00) & 0.016(0.00) & 0.025(0.00) \cr
RNNM           & 0.087(0.00) & 1.126(0.00) & 4.516(0.00) & 0.046(0.00) & 0.18(0.00) & 0.065(0.00)  \cr
LILI          & \textbf{0.005(0.01)} & \textbf{0.070(0.05)} & 2.047(0.12) & \textbf{0.013(0.01)} & \textbf{0.007(0.01)} & \textbf{0.010(0.01)} \cr
 \bottomrule[1.5pt]
\end{tabular}}
\end{table*}



Among these three tables, LILI clustering demonstrates the best performance across most datasets. Although it does not achieve the minimum loss on IHDP3, IHDP4, and ACIC16-2, it maintains a sufficiently high accuracy and its performance closely approaches that of the top methods for these datasets. Even with a number of random factors, including randomnesses in \textit{Data sampling} and \textit{Random-split}, LILI exhibits strong robustness, with its standard deviation primarily falling within the range of 0.01 to 0.03.

On the other hand, focusing on the comparison among tree-based methods, such as Causal Forest and Uplift Forest, LILI clustering consistently demonstrates its superiority across nearly all datasets. For instance, on ACIC19-3, while LILI exhibits a relatively high $L_1$ loss of 2.047, Causal Forest and Uplift Forest show significantly larger losses of 7.077 and 9.447, respectively. This highlights the efficiency of the limit inferior operation on leaf nodes.

\begin{figure*}
\centering
\includegraphics[scale=0.25]{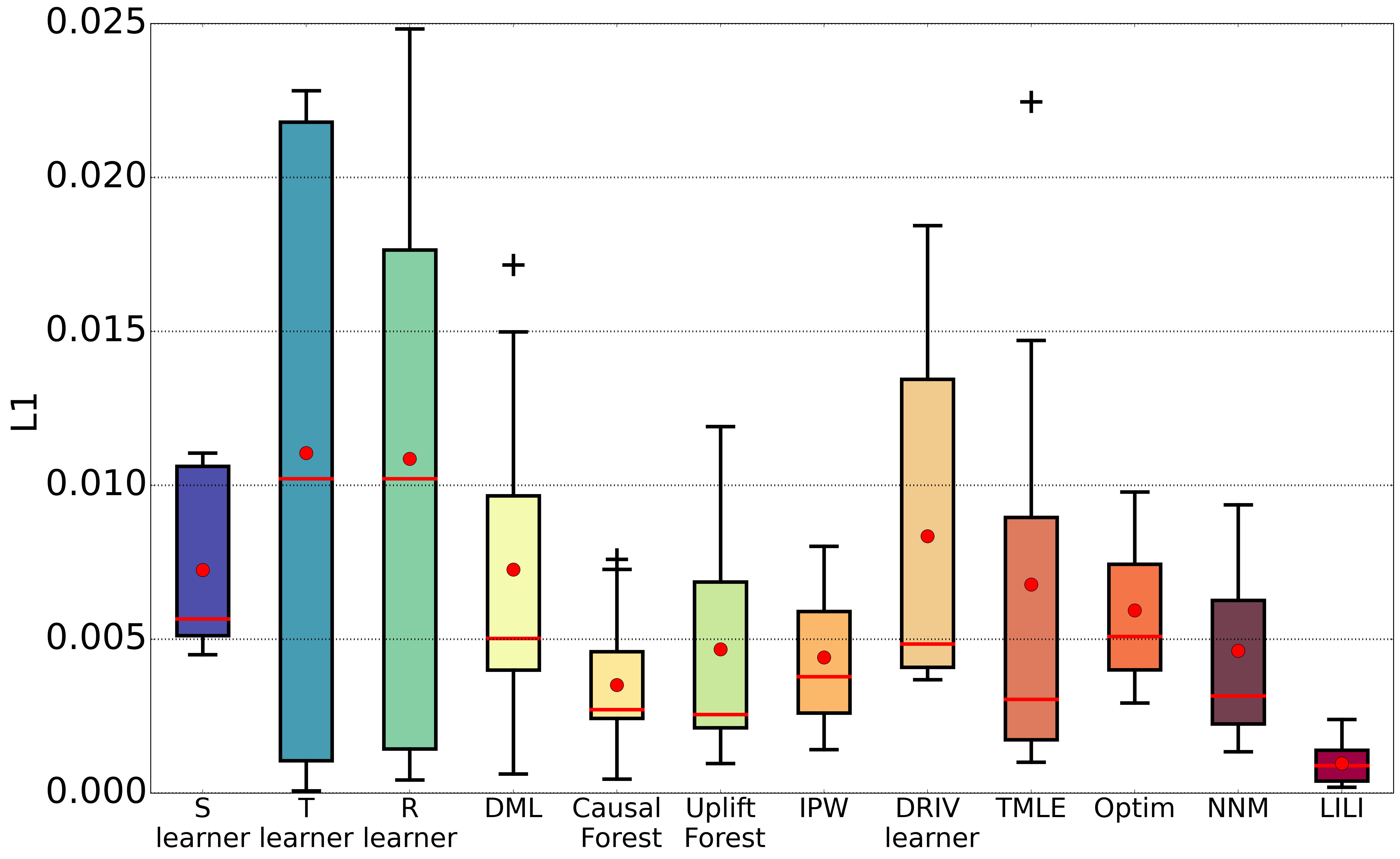}
\caption{$L_1$ losses of ATE on TWINS datasets}\label{TWINS-ATE}
\end{figure*}



For TWINS, as described in the data processing, we established a weight gap to exclude instances where twins had a weight difference below the threshold, ensuring the inclusion of meaningful variations in weight and mortality. Figure \ref{TWINS-ATE} presents a box plot comparing the performance of various methods across three weight gaps: 500 mg, 700 mg, and 900 mg. Each algorithm was run five times, resulting in 15 outcomes per method. In the box plot, the black lines indicate the maximum and minimum $L_1$ losses, while the top and bottom edges of the box represent the third and first quartiles, respectively. The red line and dot within the box denote the median and average values, with outliers marked as “+.”

From this box plot, the efficiency of LILI clustering is clear, marked by the lowest median and average values, while its robustness is underscored by the narrowest box. Other tree-based methods also demonstrate lower deviation and $L_1$ losses compared to other baseline algorithms. While re-weighting methods generally achieve higher accuracy than matching methods in terms of median and average values, their performance is inconsistent, with outliers appearing in TMLE results. In contrast, meta-learning algorithms, particularly the T-learner and R-learner, consistently exhibit the poorest performance in both accuracy and robustness. Detailed results are presented in Table \ref{L1-TWINS} and further elaborated in \textbf{Appendix B}.

\subsection{Results On ITE}


In addition to ATE, accurately predicting ITE (Individual Treatment Effect) is also crucial for causal inference algorithms, as understanding counterfactual outcomes for specific instances is often essential in real-world scenarios. In \textbf{Section 3}, we showcase the capability of LILI clustering in predicting ITE. During the clustering process, LILI clustering may exclude samples that belong to clusters with only one treatment value. To handle this, we estimate the ITE for these excluded samples using the estimated ATE.

\begin{align*}
\widehat{ITE}_i = 
\left\{
\begin{aligned}
Y_i -  \frac{1}{|I_c(\mathbf x, \Theta_K)|}\sum_{j\in I_c(\mathbf x, \Theta_K)} Y_j, \ &\text{if $i$ is saved and } T_i = 1;   \\
\frac{1}{|I_t(\mathbf x, \Theta_K)|}\sum_{j\in I_t(\mathbf x, \Theta_K)} Y_j - Y_i, \ &\text{if $i$ is saved and } T_i = 0; \\
\widehat{ATE} \quad\quad\quad\quad, \ &\text{if $i$ is abandoned}.
\end{aligned}
\right.
\end{align*}

To measure the performance of algorithms on ITE, we use the Precision in Estimation of Heterogeneous Effect (PEHE) loss, defined as 
\begin{align*}
    PEHE := \frac{1}{n}\sum^n_{i=1}\left((Y_{i1} - Y_{i0})-\widehat{ITE}_i\right)^2,
\end{align*}
where $Y_{i1}, Y_{i0}$ are the ground truth label with $T=1, T=0$. Since ACIC datasets do not provide individual counterfactual outcomes for instances, our tests focus on the IHDP and TWINS datasets.

\begin{figure*}
\centering
\includegraphics[scale=0.25]{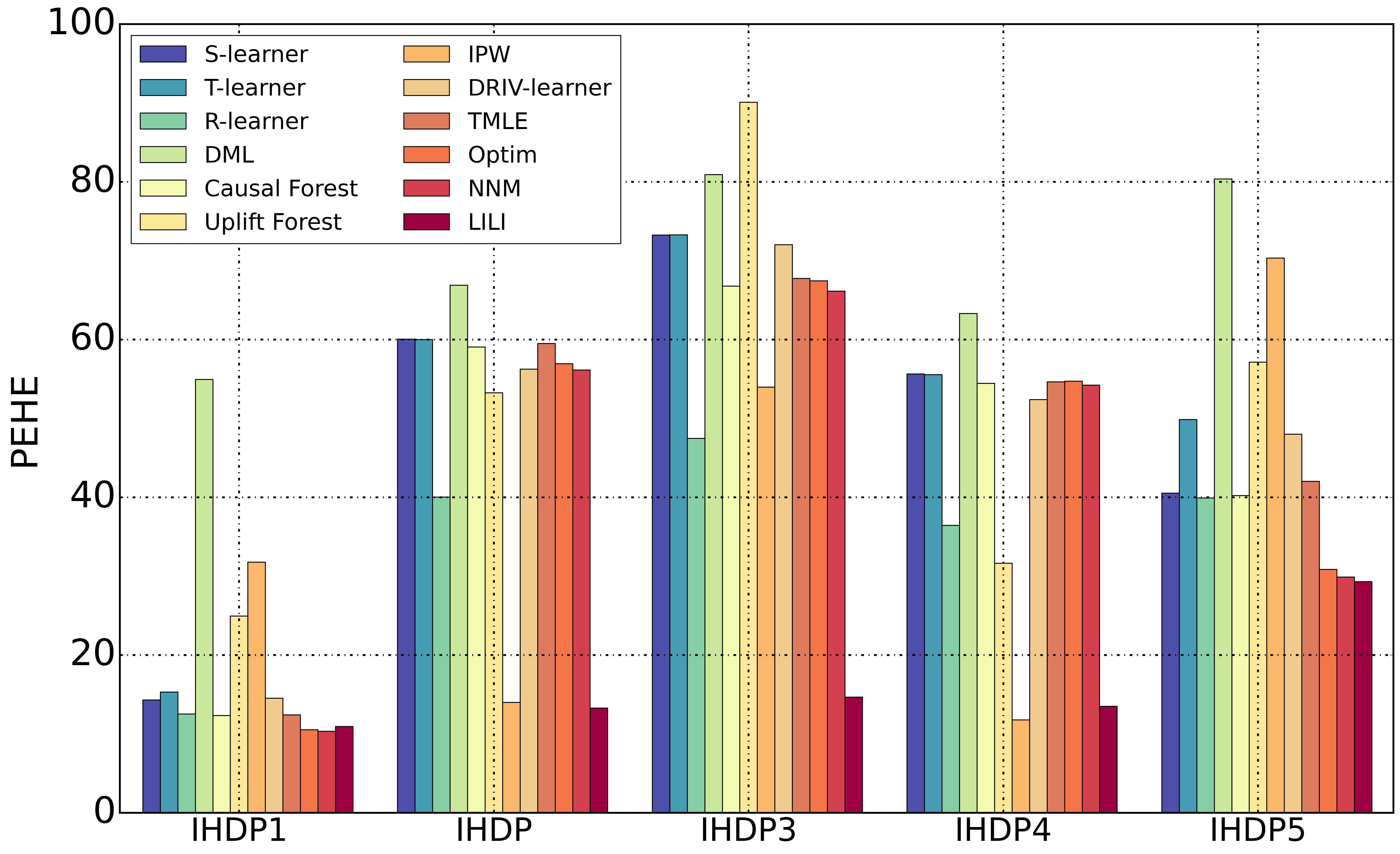}
\caption{PEHE losses of ITE on IHDP datasets}\label{bar-IHDP}
\end{figure*}



We present the PEHE results for the IHDP dataset in a bar graph in Figure \ref{bar-IHDP}. On IHDP1 and IHDP5, the proposed algorithm performs similarly to matching methods, surpassing other types of causal inference methods. On IHDP2 and IHDP4, the results of LILI are very close to those of IPW, demonstrating significantly more accurate predictions than other baseline algorithms. Notably, LILI clustering shows clear dominance on IHDP3.

\begin{figure*}
\centering
\includegraphics[scale=0.25]{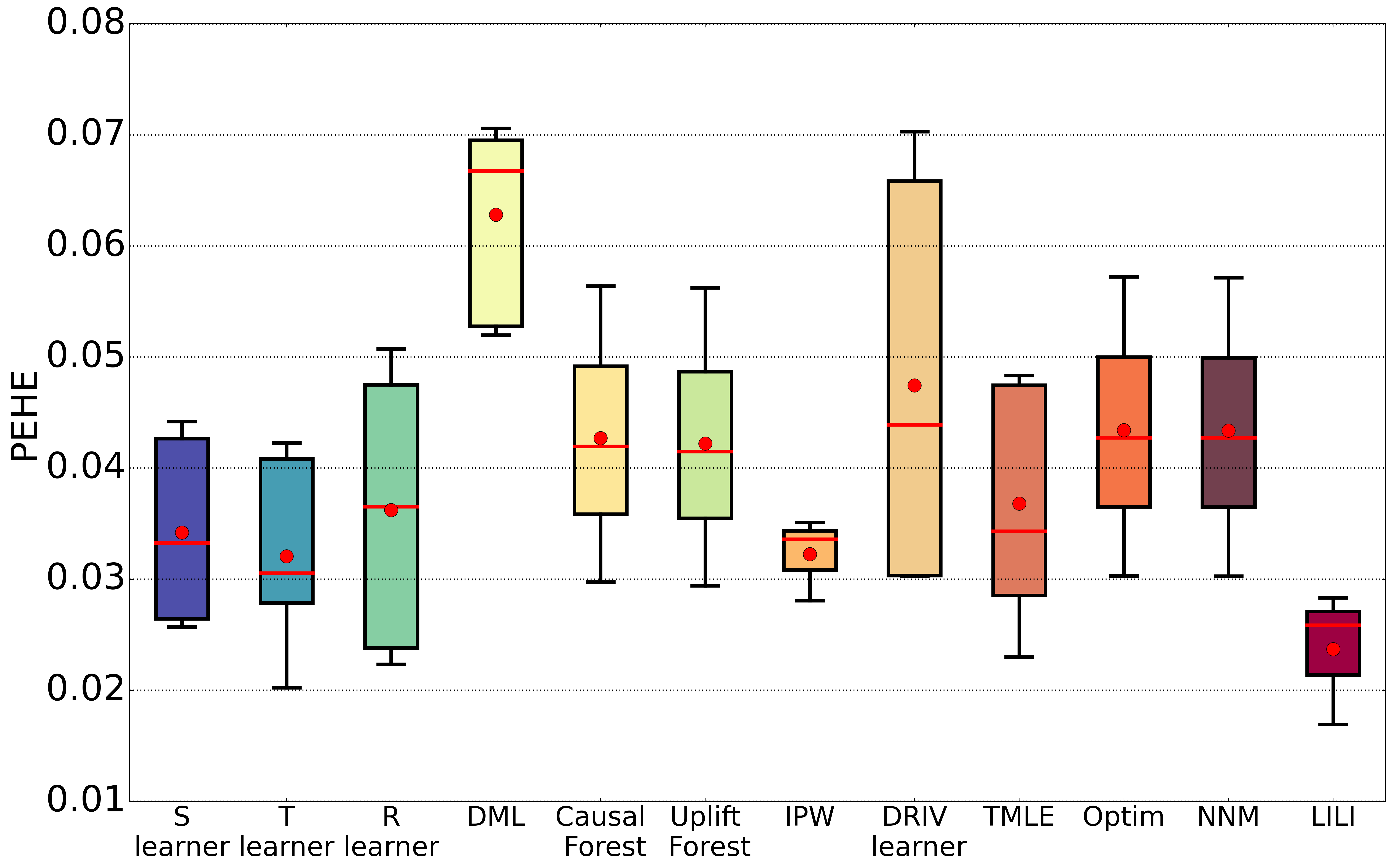}
\caption{PEHE losses of ITE on TWINS datasets}\label{TWINs-ITE}
\end{figure*}

Figure \ref{TWINs-ITE} presents the ITE prediction results on the TWINS dataset with the same three gaps. Like the ATE condition, LILI clustering exhibits the lowest median and mean values for PEHE. Tree-based methods perform better than matching methods in this context. Surprisingly, in contrast to the results in Figure \ref{TWINS-ATE}, meta-learning methods show greater accuracy than other baselines.

In conclusion, LILI clustering surpasses commonly used causal inference algorithms in predicting both the ATE and ITE. As it is based on the same causal trees, the comparison between LILI clustering and other tree-based methods highlights the efficiency of the limit inferior operation at leaf nodes. In addition to high accuracy, LILI clustering demonstrates strong robustness, indicated by small deviation values across experimental datasets. The precise mean and standard deviation values for these two figures are presented in Tables \ref{PEHE-IHDP}, \ref{PEHE-TWINS} in \textbf{Appendix B}.
We will now discuss strategies for fine-tuning the hyperparameters in LILI.

\subsection{Hyperparameters analyses}

In experiments, most of the hyperparameters of proposed methods are fixed. The adjustable hyperparameters are those mentioned in Table \ref{data-info}. However, due to the context of \textit{$\alpha$-regularity}, the parameter $\alpha$ is constrained by $l$, i.e. 
\begin{align*}
    \frac{l}{n} \leq \alpha \leq \frac{2l-1}{n},
\end{align*}
and in experiments, we set $\alpha = l/n$. Therefore, $l$ and $K$ are critical hyperparameters that significantly impact the prediction of ATE. Fine-tuning these two parameters is essential.


Under limited samples, $l$ and $K$ influence the same factors: the number of clusters $|\mathcal I|$ and the number of abandoned instances. Smaller $l$ and larger $K$ impose stricter matching requirements, which can improve the accuracy of counterfactual outcome estimates for each LILI. However, these stricter requirements also reduce the likelihood of finding suitable instances for matching, leading to an increase in the number of clusters while simultaneously decreasing the number of available instances. Fine-tuning causal inference methods is more challenging than traditional machine learning algorithms due to the lack of ground truth labels in real-world scenarios, where training and validation datasets are often unavailable. In the context of LILI clustering, the number of clusters and available instances are observable, providing metrics that can aid in the fine-tuning process.

\begin{figure*}
\centering

\subfigure[]
{
    \begin{minipage}[b]{.48\linewidth}
        \centering
        \includegraphics[scale=0.34]{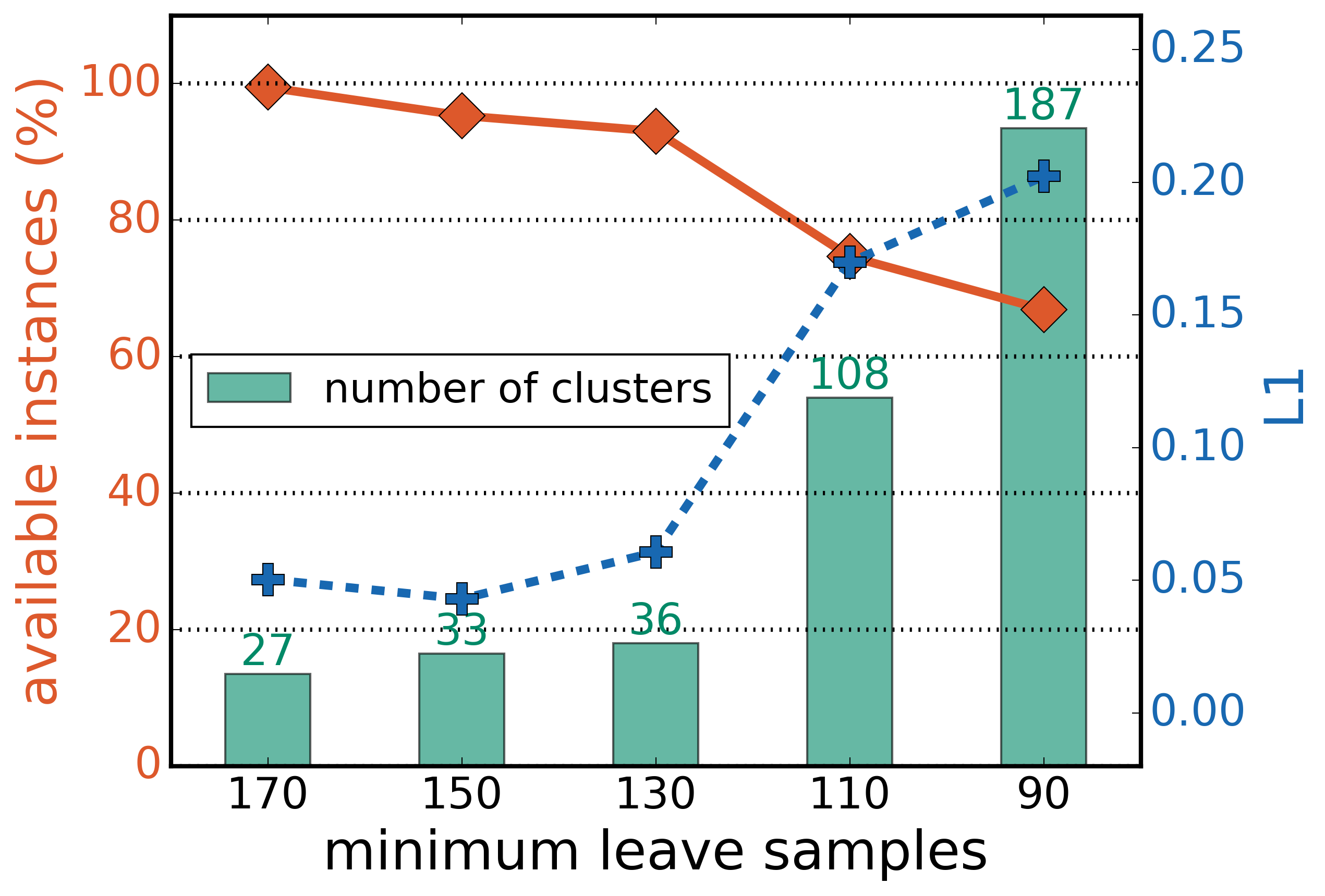}
    \end{minipage}
}
\subfigure[]
{
    \begin{minipage}[b]{.48\linewidth}
        \centering
        \includegraphics[scale=0.34]{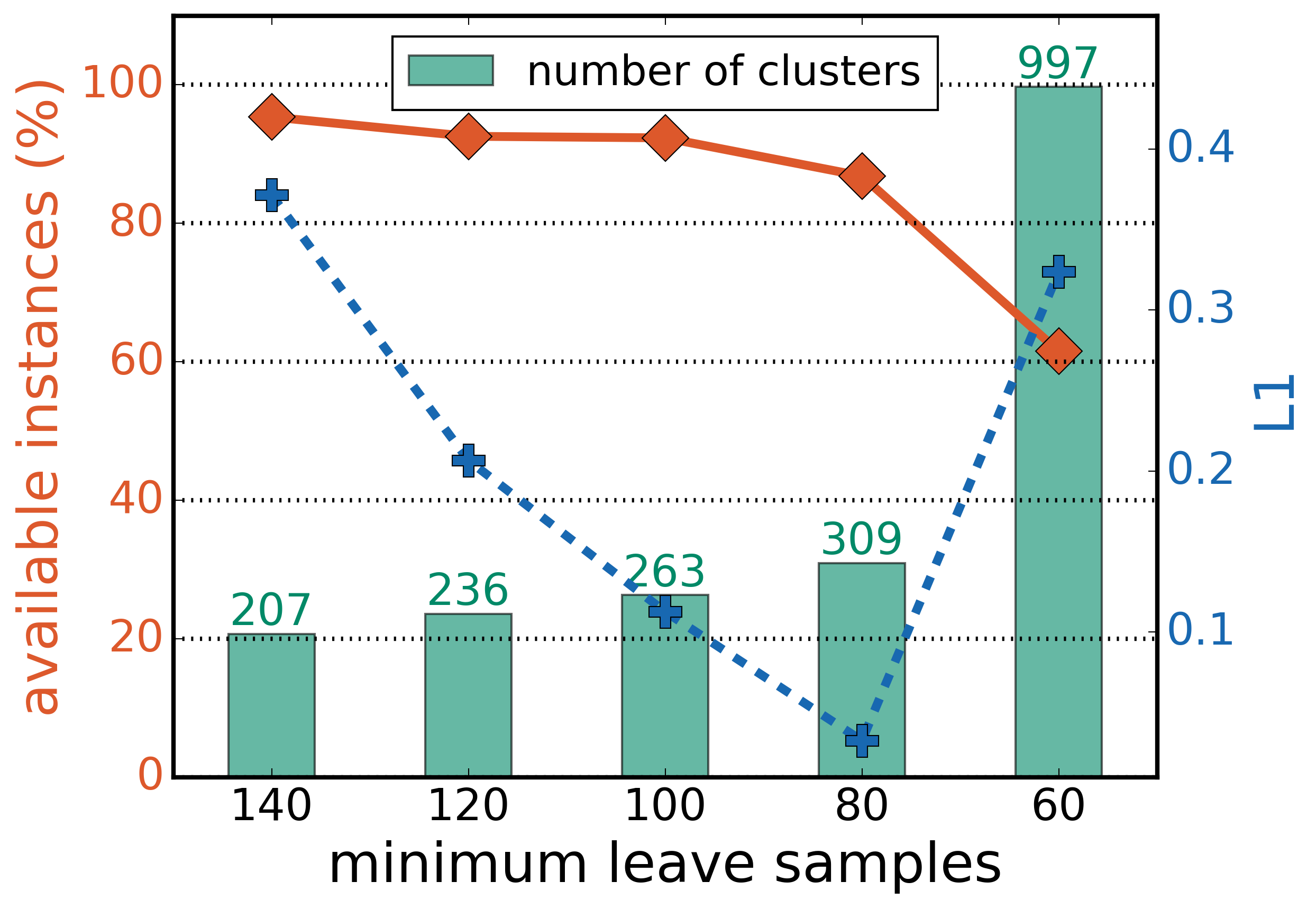}
    \end{minipage}
}
\subfigure[]
{
    \begin{minipage}[b]{.48\linewidth}
        \centering
        \includegraphics[scale=0.34]{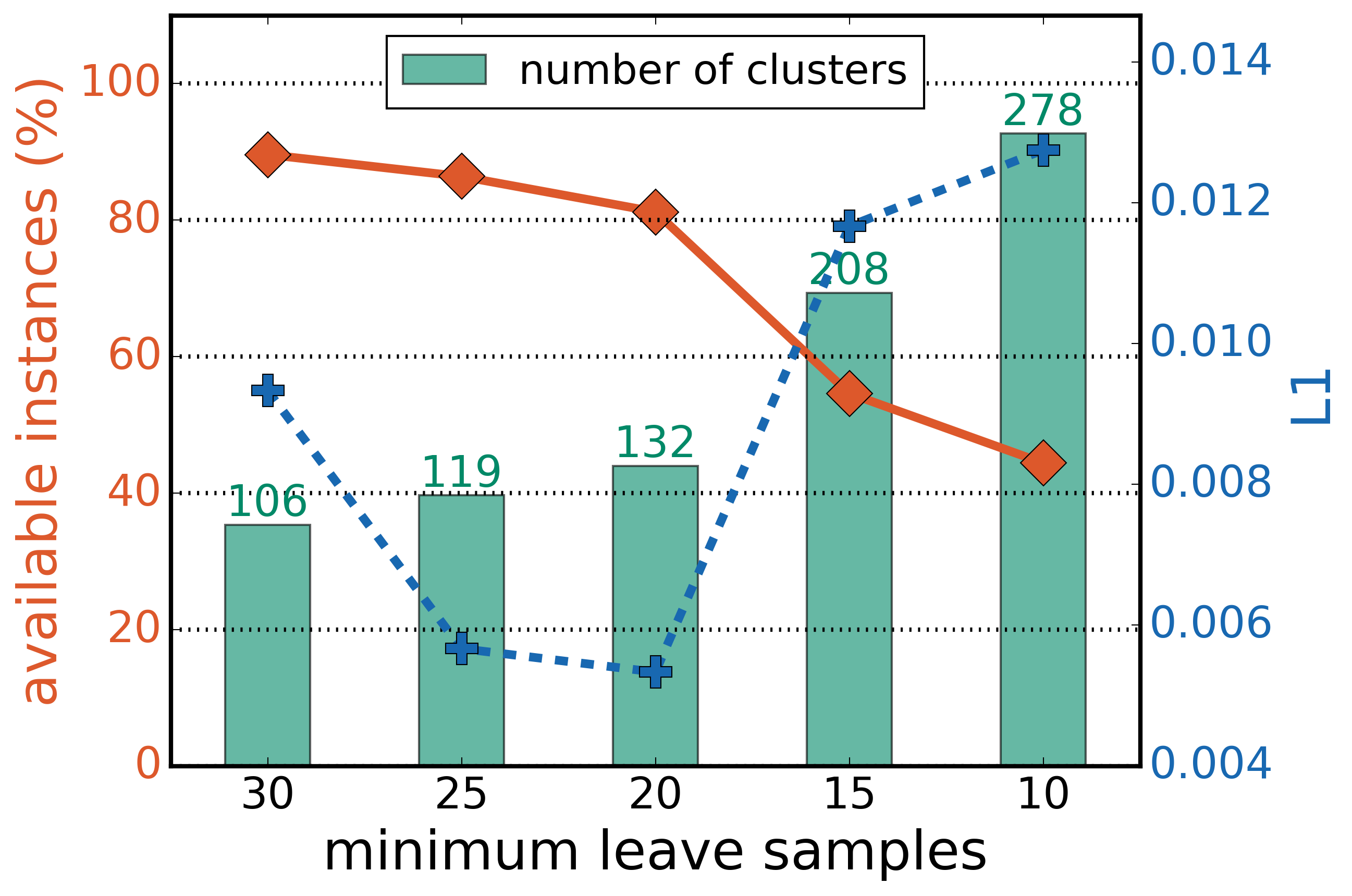}
    \end{minipage}
}
\subfigure[]
{
    \begin{minipage}[b]{.48\linewidth}
        \centering
        \includegraphics[scale=0.34]{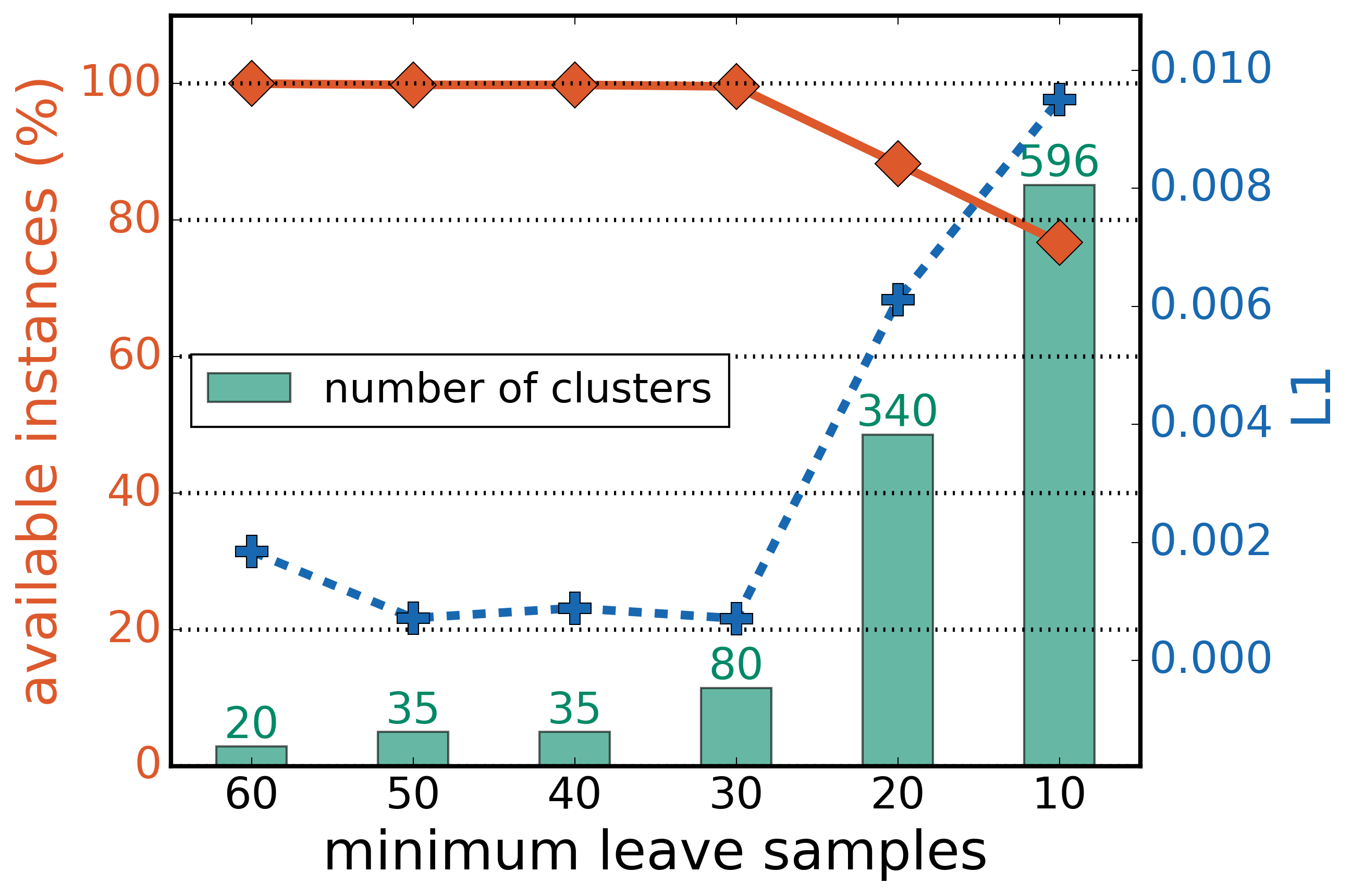}
    \end{minipage}
}

\caption{(a), (b), (c) and (d) present the results of the LILI clustering algorithms under varying minimum leaf samples $l$ on the IHDP1, ACIC16-1, ACIC19-1 and TWINS datasets. The red solid lines represent the percentage of available instances, the blue dotted lines indicate the $L_1$ losses, and the green bars display the number of clusters as $l$ changes. The x-axis denotes the value of $l$, while the left y-axis and right y-axis reflect the percent of available instances and $L_1$ losses. The values for the number of clusters are directly marked at the top of the bars.}\label{hyper-L}
\end{figure*}

Figure \ref{hyper-L} presents an analysis of LILI clustering with respect to minimum leaf samples $l$, while keeping $K$ fixed at the corresponding value listed in Table \ref{data-info}. As expected, across all graphs, the number of available instances decreases as $l$ decreases, while the number of clusters increases. Several common patterns emerge across these four graphs: with a decrease in $l$, the available instances initially remain relatively stable but then experience a significant drop at certain points. Concurrently, the number of clusters increases dramatically. The $L_1$ loss for ATE decreases or slightly fluctuates initially but rises sharply following these changes in the two metrics. For example, in Figure \ref{hyper-L} (c) ACIC19-1, as $l$ decreases from 30 to 20, available instances decline gradually from 90\% to 80\%, while the number of clusters stays around 100-150. The $L_1$ loss decreases initially but stabilizes at $l=25$. However, when $l$ drops to 15, available instances decrease to approximately 50\%, and the number of clusters increases to 208, causing a sharp rise in $L_1$ loss. Thus, the optimal value for $l$ is just before the sudden change in the number of available instances and clusters. In this case, the best value for $l$ is 20.

Different from $K$ in the original causal forest, a larger $K$ in LILI clustering may not necessarily yield better results.  The function $f(K) = \sqrt{K}$ suggests that the requirements for gathering two instances into the same LILI become stricter as $K$ increases, since $K-\sqrt{K}$ is an increasing function. Consequently, an increase in causal trees can lead to more clusters and fewer available instances, indicating that similar considerations apply when fine-tuning $K$. 

\begin{figure*}
\centering

\subfigure[]
{
    \begin{minipage}[b]{.48\linewidth}
        \centering
        \includegraphics[scale=0.15]{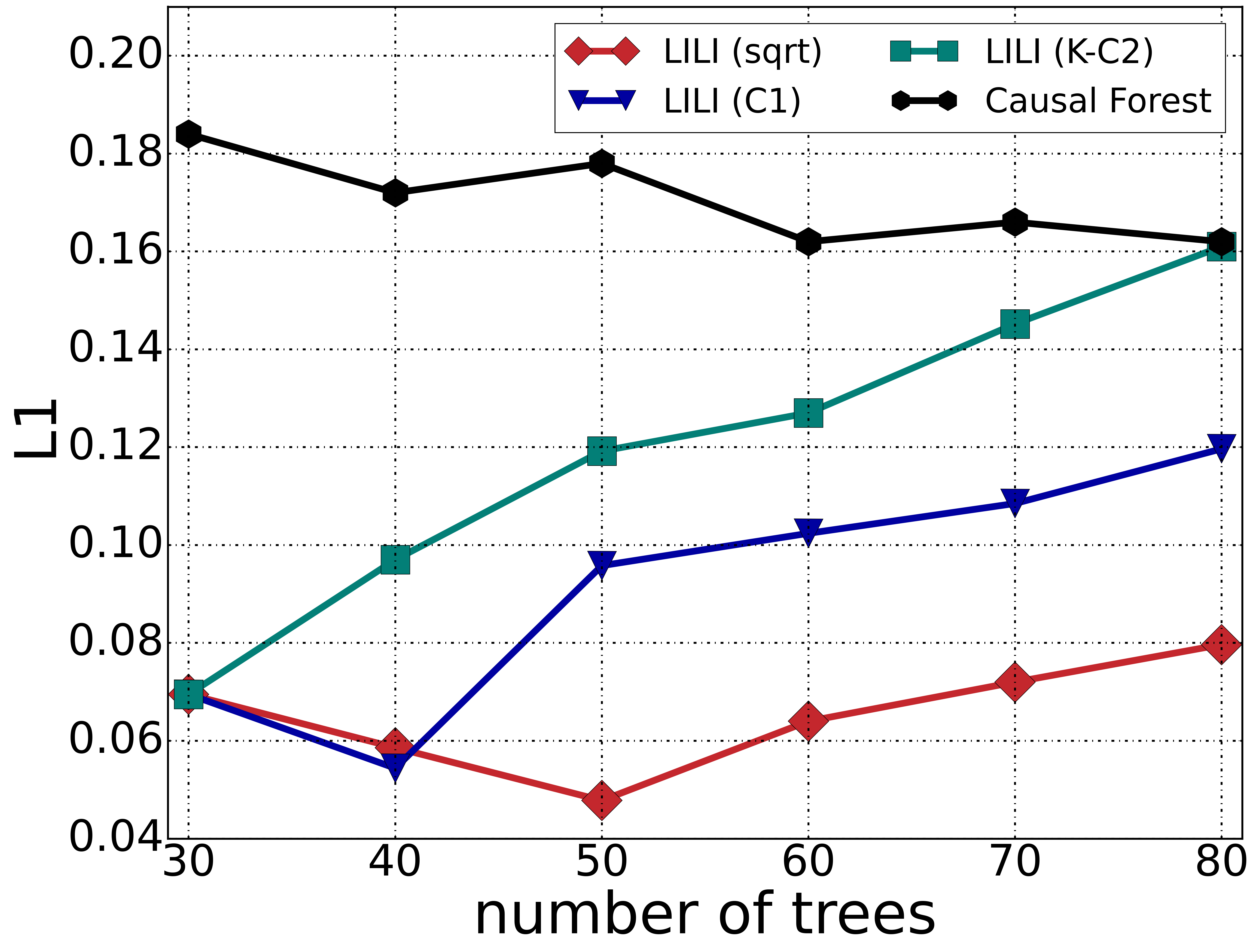}
    \end{minipage}
}
\subfigure[]
{
    \begin{minipage}[b]{.48\linewidth}
        \centering
        \includegraphics[scale=0.15]{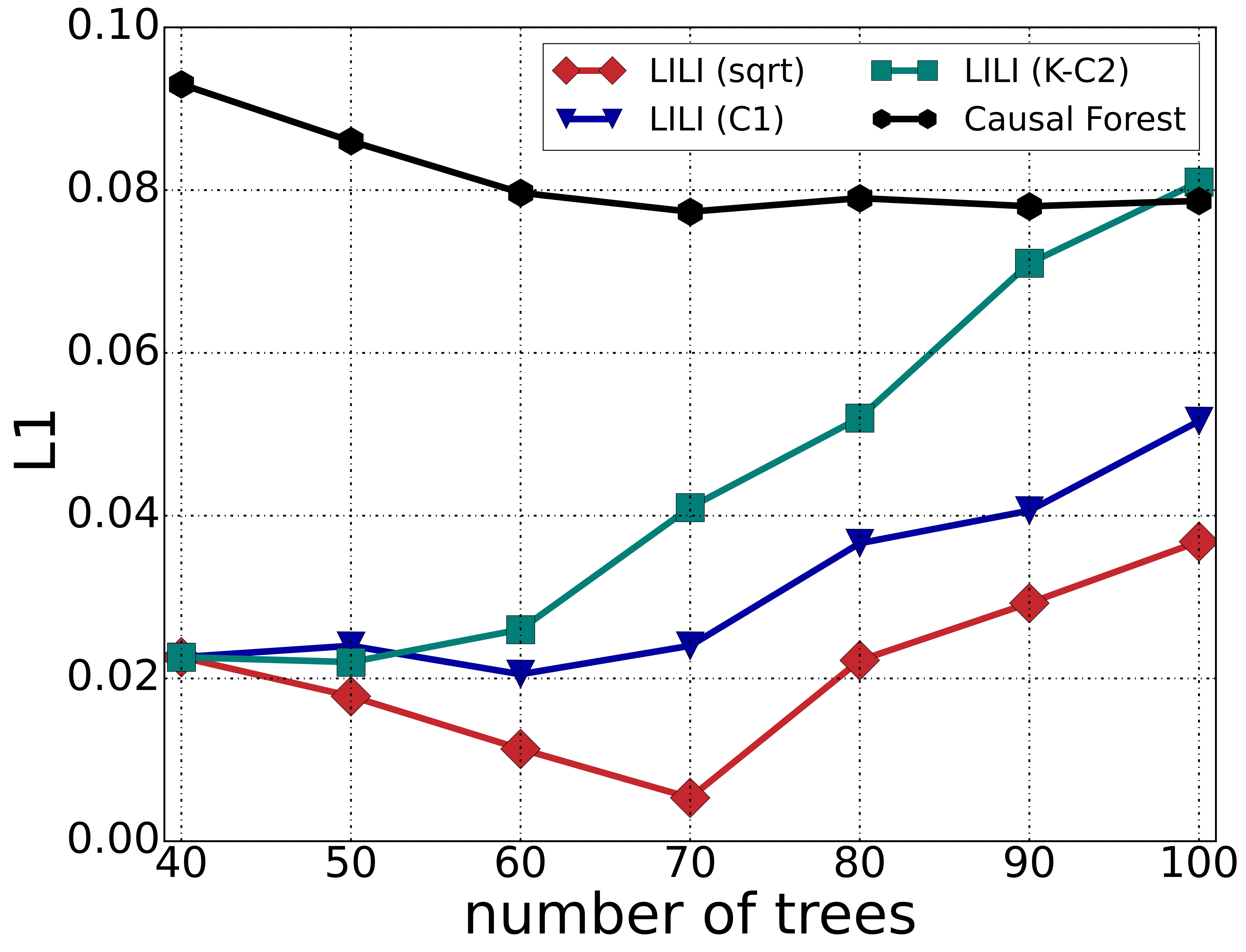}
    \end{minipage}
}

\caption{The $L_1$ loss on ATE with respect to the number of trees $K$. (a) and (b) present the results of causal forest and LILI clustering with various tolerance functions on IHDP1 and ACIC19-1 datasets, respectively. The LILI variants include LILI (sqrt), LILI (C1) and LILI (K-C2), corresponding to the tolerance functions $\sqrt{K}, C_1$ and $K-C_2$ where $C_1, C_2$ are constants, specifically $C_1=\sqrt{40}$, $C_2 = 40-\sqrt{40}$ in (a) and $C_1=\sqrt{30}$, $C_2 = 30-\sqrt{30}$ in (b)}\label{hyper-K}
\end{figure*}

Recall from \textbf{Section 4} that we provide two ranges for tolerance function: the allowed range in \textbf{Theorem \ref{LILI-extention1}} and the forbidden range in \textbf{Theorem \ref{LILI-extention2}}. Figure \ref{hyper-K} empirically validates these two theorems by selecting $f(K) = C_1$ from (\ref{f-require1}) and $f(K) = K- C_2$ from (\ref{f-require2}). To facilitate a better comparison between LILI clustering methods with different tolerance functions, we adjust the values of $C_1, C_2$ so that they all start from the same point. 


Within these two graphs, the $L_1$ loss for the original LILI clustering decreases initially and then rises with the growth of \( K \). This pattern mirrors the behavior seen in Figure \ref{hyper-L}. LILI (C1) has a more restrictive tolerance function than LILI (\(\sqrt{K}\)), leading to stricter clustering requirements. While it initially performs similarly or even better than LILI (\(\sqrt{K}\)), its performance degrades as \( K \) increases. For LILI (K-C2), as analyzed in \textbf{Theorem \ref{LILI-extention2}}, bias in LILI increases with the number of causal trees. Consequently, increasing \( K \) has a negative impact on prediction, and the $L_1$ losses for LILI (K-C2) continue to rise, showing significantly poorer effectiveness than other LILI clustering algorithms and ultimately approaching the performance of causal forests.

Overall, the available instances and number of clusters are the main indices for fine-tuning the hyperparameters. In experiments, we can use traditional grid search to identify the optimal values $l$ and $K$. The best combination of $l$ and $K$ ensures that
Overall, the number of available instances and the number of clusters are the primary indices for fine-tuning the hyperparameters. In experiments, traditional grid search can be employed to identify the optimal values for \( l \) and \( K \). The best combination of \( l \) and \( K \) ensures that 
\begin{itemize}
    \item The percent of available instances in LILI clustering does not fall below 90\% or 80\%;
    \item LILI clustering with $l,K$ maximizes the number of clusters while satisfying the first condition.
\end{itemize}
The first condition is derived from Figure \ref{hyper-L}. The available instances at sudden change points are consistently above 80\%. The second condition aims to find the optimal values of \(l\) and \(K\) from the model that meet the first condition. If the dataset size or model limitations restrict the complexity of the grid search, we will prioritize maintaining a high number of available instances by decreasing \(l\) and increasing \(K\) as much as possible while still adhering to the first condition.

Most surprisingly, the number of causal trees does not always positively impact performance in LILI clustering. This represents a key difference from other tree-based methods. Since experiments cannot always obtain unlimited samples, it also emphasizes that in the definition of LILI, the limit \(n \to \infty\) must be considered before \(K \to \infty\).


\section{Conclusion}
To eliminate potential biases and enhance the accuracy of the causal tree, we aim to identify all instances $\mathbf y$ for which $P(\mathbf y\in L(\mathbf x))=1$ to match $\mathbf x$ for the causal inference task. By constructing LILI $I(\mathbf x)$, this question can be transformed into finding $\mathbf y$ such that $P(\mathbf y\in I(\mathbf x))=1$. Using the connection between LILI, LSLI, and limit inferior and superior in set theory, we propose the zero-one law for LILI and LSLI, which allows us to derive $\mathbf y$ with $P(\mathbf y\in L(\mathbf x))=1$ from $\mathbf y\in I(\mathbf x)$. Furthermore, we introduce the proposed algorithm: LILI clustering, and prove the convergence of estimated ATE from LILI clustering by demonstrating the convergence of estimated distribution induced by LILI. We then explore the permitted and forbidden areas for adjusting the tolerance function in LILI. Finally, we empirically validate the efficiency of LILI clustering compared to other baselines on common datasets and provide analyses about two key hyperparameters. 

Despite the notable symmetry and superior performance on LILI clustering, there are still certain limitations. For every pair of instances, constructing LILI requires counting the number of trees that classify them into the same leaves, making this procedure significantly more complex and time-consuming than causal forest. Additionally, LILI clustering excludes some instances to enhance ATE predictions, but it neglects the precise prediction of ITEs for these excluded samples. Therefore, future work will focus on optimizing LILI construction and developing methods to predict ITEs for these instances.

\acks{This work was funded by the National Nature Science Foundation of China under Grant 12320101001.}


\newpage

\appendix
\section{}
\label{app:theorem}



In this appendix, we provide all proofs of Lemmas, Propositions, Theorems and Corollaries appearing in \textbf{Section 3} and \textbf{Section 4}.\\~

\begin{proof}{\textbf{of Lemma \ref{lb-cf}}}\\
Let $c(\mathbf x)$ denote the depth of the leaf node containing $\mathbf x$. As the \textit{$\alpha$-regularity}, every split node to the $L(\mathbf x)$ at least contains $\alpha s_n$ samples. If the child of a split node is still a split node, it also contains at least $\alpha s_n$ samples. But it contains at most $(1-\alpha)s_n$, otherwise, the other child node of its parent contains instances less than $\alpha s_n$ and this breaks the \textit{$\alpha$-regularity}. After $c(\mathbf x)-1$ and $c(\mathbf x)$ times of splits, as the node on the level $c(\mathbf x)-1$ is still a split node, we have 
\begin{align*}
\alpha^{c(\mathbf x)}s_n \leq 2l-1, \\
(1-\alpha)^{c(\mathbf x)-1} s_n \geq \alpha s_n.
\end{align*}
Therefore, the depth of $L(x)$ has an upper bound $ \frac{\ln\alpha}{\ln(1-\alpha)}+1$. Since $c(\mathbf x)$ is an integer, its upper bound can be written as $\lceil\frac{\ln\alpha}{\ln(1-\alpha)}\rceil$. Since the two instances $\mathbf x, \mathbf y$ have the same value in the first feature, if we only pick $X_1$ as the split feature, we can never split $\mathbf x, \mathbf y$. By the \textit{Random-split} property, on every split node, we pick $X_1$ as the split feature under at least $\pi/d$ probability. Since $L(\mathbf x)$ is on the level of $c(\mathbf x)$, it goes through $c(\mathbf x)$ split nodes. Therefore, the probability that all the split nodes on the road to $L(\mathbf x)$ is $X_1$ has a lower bound:
\begin{align*}
    \left(\frac{\pi}{d}\right)^{\left\lceil\frac{\ln\alpha}{\ln(1-\alpha)}\right\rceil}\leq \left(\frac{\pi}{d}\right)^{c(\mathbf x)} \leq P(\mathbf y\in L(\mathbf x)).
\end{align*}
\end{proof}

\begin{proof}{\textbf{of Proposition \ref{LSLI}}} \\
Let $\mathbf y$ be another instance and $\mathbf y \in I^c(\mathbf x, \Theta_K)$. Then $\mathbf y$ must not belong to any of the intersection of leaf intervals $\bigcap_{k=1}^s I(\mathbf x, \theta_{i_k})$ when $s>K-\sqrt{K}$, i.e., any $s>K-\sqrt{K}$ trees at least contain one tree that does not assign $\mathbf x$ and $\mathbf y$ into one leaf. Therefore, we can find at least $m = \lceil\sqrt{K}\rceil$ causal trees $\theta_{j_1}, \theta_{j_2}, ... \theta_{j_m}$ such that $\mathbf y \notin I(\mathbf x, \theta_{j_i})$ for $i=1,2,...,m$. For any $s>K-\sqrt{K}$ trees, there is a $j_a\in \{i_1, i_2, ..., i_s\}$ then $I^c(\mathbf x, \theta_{j_a})$ contains $\mathbf y$ for some $a$. Therefore, 
\begin{align*}
    \mathbf y\in \bigcup_{k=1}^s I^c(\mathbf x, \theta_{i_k}),
\end{align*}
for any $s>K-\sqrt{K}$. It means that $\mathbf y\in \bar I^c(\mathbf x, \Theta_K)$ and 
\begin{align*}
    I^c(\mathbf x, \Theta_K) \subseteq \bigcap_{1\leq i_1<\cdots i_s\leq K\atop s>K-\sqrt{K}}\bigcup_{k=1}^{s} I^c(\mathbf x, \theta_{i_k}).
\end{align*}
On the other hand, if $\mathbf y\in\bar I^c(\mathbf x, \Theta_K)$, then for any $s>K-\sqrt{K}$, $\mathbf y\in \bigcup^s_{k=1} I^c(\mathbf x, \theta_{i_k})$, and we can at least find one tree $i_a$ such that split $\mathbf x$ and $\mathbf y$ into different leaves. We traverse all possible $s$ and gather all this tree as $\{j_1, j_2, ..., j_m\}$. We assert that $m\geq\lceil \sqrt{K}\rceil$, otherwise, the union of the rest of the trees $\{i_1, i_2, ..., i_{K-m}\}$ where $K-m >K-\sqrt{K}$ satisfies $\mathbf y\notin \bigcup_{k=1}^{K-m} I(\mathbf x, \theta_k)$ then $\mathbf y\notin \bar I^c(\mathbf x, \Theta_K)$ which contradicts the requirement. Therefore, for any $s>K-\sqrt{K}$, there is a $j_a$, $j_a\in \{i_1, i_2, ..., i_s\}$ and $\mathbf y\notin \bigcap^s_{k=1} I(\mathbf x, \Theta_K)$, i.e. $\mathbf y\in I^c(\mathbf x, \Theta_K)$. Then we have 
\begin{align*}
    I^c(\mathbf x, \Theta_K) \supseteq \bigcap_{1\leq i_1<\cdots i_s\leq K\atop s>K-\sqrt{K}}\bigcup_{k=1}^{s} I^c(\mathbf x, \theta_{i_k}).
\end{align*}
And $I^c(\mathbf x, K) = \bar I^c(\mathbf x, \Theta_K)$ for any $K$. When $K\to\infty$ and $n\to\infty$, the equality still holds. This completes the proof.
\end{proof}

\begin{proof}{\textbf{of Proposition \ref{LILI-LSLI}}}\\
For (1), if $\mathbf y\in I(\mathbf x, \Theta_K)$, we can find a set of causal trees $\{i_1, i_2, ..., i_m\}$ where $m>K-\sqrt{K}$ such that all the trees in this set classify $\mathbf x$ and $\mathbf y$ into the same leaf. Then for any union of leave intervals on $s$ trees where $s>K-\sqrt{K}$, they must contain at least one tree in $\{i_1, i_2, ..., i_m\}$, so
\begin{align*}
    \mathbf y \in \bigcup^s_{k=1} I(\mathbf x, \theta_{i_k}).
\end{align*}
Therefore, we have $I(\mathbf x, \Theta_K)\subseteq \bar I(\mathbf x, \Theta_K)$. The conclusion achieves for any $K$ and also achieves for $n,K\to\infty$. 

For (2), since the monotonicity of $I(\mathbf x, \theta_i)$, we have
\begin{align*}
    I(\mathbf x) = \lim_{K\to\infty}\lim_{n\to\infty} \bigcup_{1\leq i_1<\cdots i_s\leq K\atop s>K-\sqrt{K}}\bigcap_{k=1}^{s} I(\mathbf x, \theta_{i_k}) = \lim_{K\to\infty}\lim_{n\to\infty} \bigcup_{1\leq i_1<\cdots i_s\leq K\atop s>K-\sqrt{K}} I(\mathbf x, \theta_{i_1}) = \lim_{K\to\infty}\lim_{n\to\infty}\bigcup_{k=1}^{\lceil\sqrt{K}\rceil} I(\mathbf x, \theta_k).
\end{align*}
In the last step, the intersection of the leaf intervals is equal to the interval that has the smallest index. As we need to traverse all the subsets of causal trees with $s>K-\sqrt{K}$, the smallest index in any subset of causal trees must belong to $\{1, 2, ..., \lceil\sqrt{K}\rceil\}$. In the conclusion, we have $I(\mathbf x) = \lim_{K\to\infty}\lim_{n\to\infty} \bigcup_{k=1}^{\lceil\sqrt{K}\rceil} I(\mathbf x, \theta_k) = \bigcup^\infty_{k=1} I(\mathbf x, \theta_k)$. For $\bar I(\mathbf x)$, the union of the first $s$ leaf intervals is dominant. So we can obtain
\begin{align*}
    \bar I(\mathbf x) = \lim_{K\to\infty}\lim_{n\to\infty} \bigcap_{1\leq i_1<\cdots i_s\leq K\atop s>K-\sqrt{K}}\bigcup_{k=1}^{s} I(\mathbf x, \theta_{i_k}) = \lim_{K\to\infty}\lim_{n\to\infty} \bigcup^{K-\sqrt{K}}_{k=1}I(\mathbf x, \theta_k) = \bigcup^\infty_{k=1}I(\mathbf x, \theta_k).
\end{align*}
The proof of (3) is similar.
\end{proof}

\begin{proof}{\textbf{of Theorem \ref{core-lemma}}}\\
If $P(\mathbf y\in I(\mathbf x))=1$, let us assume $\lim_{n\to\infty} P(\mathbf y\in L(\mathbf x))=p<1$. Since the constructions of causal trees are independent, $P(\mathbf y\in I(\mathbf x, \theta_i)\bigcap I(\mathbf x, \theta_j)) = P(\mathbf y\in I(\mathbf x, \theta_i))P(\mathbf y\in I(\mathbf x, \theta_j))$, then for any $s>K-\sqrt{K}$, 
\begin{align*}
    \lim_{n\to\infty}P\left(\mathbf y\in\bigcap_{k=1}^s L_{i_k}(\mathbf x)\right) = \lim_{n\to\infty}P\left(\mathbf y\in\bigcap_{k=1}^s I(\mathbf x, \theta_{i_k} )\right) = \lim_{n\to\infty}\prod_{k=1}^s P\left(y\in I(\mathbf x,\theta_{i_k})\right) = p^s, 
\end{align*}
\begin{align*}
    P\left(\mathbf y\in I(\mathbf x)\right) 
    &= P\left(\mathbf y\in\lim_{K\to\infty}\lim_{n\to\infty}\bigcup_{1\leq i_1<\cdots i_s\leq K\atop s>K-\sqrt{K}}\bigcap^s_{k=1} I(\mathbf x,\theta_{i_k})\right)
    \leq \lim_{K\to\infty}\lim_{n\to\infty}\sum_{1\leq i_1<\cdots i_s\leq K\atop s>K-\sqrt{K}} P\left(\mathbf y\in\bigcap^s_{k=1} I(\mathbf x,\theta_{i_k})\right) \\
    &\leq \lim_{K\to\infty}\lim_{n\to\infty}\sum_{1\leq i_1<\cdots i_s\leq K\atop s>K-\sqrt{K}} P(\mathbf y\in I(\mathbf x, \theta))^s \\
    &\leq \lim_{K\to\infty}p^{K-\sqrt{K}}\left(\binom{K}{K}+\binom{K}{K-1}+\cdots \binom{K}{K-\left[\sqrt{K}\right]} \right).
\end{align*}
In the last term, we need to traverse any possible $s>K-\sqrt{K}$, the number of possible choices of $s$ is $\sum_{i=0}^{[\sqrt{K}]}\binom{K}{K-i}$ where $\binom{K}{m}$ is the number of possible choices on picking $m$ items from a total of $K$ items. When $K$ is large enough and $m< K/2$, $\binom{K}{m}$ is monotone increasing with respect to $m$. We can magnify the above equation,  
\begin{align*}
    P(\mathbf y\in I(\mathbf x)) 
    &\leq \lim_{K\to\infty} p^{K-\sqrt{K}}\sum^{\left[\sqrt{K}\right]}_{i=0}\binom{K}{K-i} \\
    &=\lim_{K\to\infty}p^{K-\sqrt{K}}\sum^{\left[\sqrt{K}\right]}_{i=0}\binom{K}{i} \\
    &\leq \lim_{K\to\infty}\lim_{n\to\infty} p^{K-\sqrt{K}}\left(1+\left[\sqrt{K}\right] \binom{K}{\left[\sqrt{K}\right]}\right) \\
    &\leq \lim_{K\to\infty} p^{K-\sqrt{K}}\left(1+K^{\sqrt{K}+1} \right).
\end{align*}
By L'Hospital's rule, when $K\to\infty$, $\left((K-\sqrt{K})\ln p+\left(\sqrt{K}+1\right)\ln K\right)/\left(K\ln p\right)\to 1$, 
\begin{align*}
p^{K-\sqrt{K}}\left(1+K^{\sqrt{K}+1} \right) = p^{K-\sqrt{K}} + \exp\left((K-\sqrt{K})\ln p+\left(\sqrt{K}+1\right)\ln K\right) = p^{K-\sqrt{K}} + \exp\left(K\ln p\right)\to 0. 
\end{align*}
Thus $P(\mathbf y\in I(\mathbf x)) = 0$ and it contradicts the requirement that $P(\mathbf y\in I(\mathbf x))=1$.

On the other hand, if $\lim_{n\to\infty}P(\mathbf y\in L(\mathbf x))=1$, 
\begin{align*}
    P(\mathbf y\in I(\mathbf x))
    &= P\left(\mathbf y\in\lim_{K\to\infty}\lim_{n\to\infty}\bigcup_{1\leq i_1<\cdots i_s\leq K\atop s>K-\sqrt{K}}\bigcap^s_{k=1} I(\mathbf x,\theta_{i_k})\right)\\
    &\geq P\left(\mathbf y\in\lim_{K\to\infty}\lim_{n\to\infty}\bigcap_{k=1}^K I(\mathbf x, \theta_k)\right)\\
    &= \lim_{K\to\infty}\lim_{n\to\infty}\prod^K_{k=1}P(\mathbf y\in I(\mathbf x, \theta_k))\\
    &= \lim_{K\to\infty}\lim_{n\to\infty}\prod^K_{k=1}P(\mathbf y\in L(\mathbf x)|\mathcal D) = 1.
\end{align*}

All the above discussions state that $P(\mathbf y\in I(\mathbf x))=1 \Longleftrightarrow\lim_{n\to\infty} P\left(\mathbf y\in L(\mathbf x)|\mathcal D\right) = 1$.
\end{proof}

\begin{proof}{\textbf{of Corollary \ref{LILI-01}}}\\
$P(\mathbf y\in L(\mathbf x))=1 \Longrightarrow P(\mathbf y\in I(\mathbf x))=1$ is trivial according to \textbf{Theorem \ref{core-lemma}}. From the proof of \textbf{Theorem \ref{core-lemma}}, we know if $P(\mathbf y\in L(\mathbf x))<1$,
\begin{align*}
    P(\mathbf y\in I(\mathbf x))\leq \lim_{K\to\infty}\lim_{n\to\infty} p^{K-\sqrt{K}}\left(1+K^{\sqrt{K}+1} \right) = 0.
\end{align*}
\end{proof}

\begin{proof}{\textbf{of Theorem \ref{LSLI-01}}}\\
Firstly, let us assume $P(\mathbf y\in L(\mathbf x))=0$.
As $\bar I(\mathbf x)\subset \bigcup_{k=1}^KI(\mathbf x, \theta_k)$, and $P(\mathbf y\in I(\mathbf x, \theta_k))=P(\mathbf y\in L(\mathbf x))$
\begin{align*}
    P(\mathbf y\in\bar I(\mathbf x)) \leq P\left(\mathbf y\in \bigcup^K_{k=1} I(\mathbf x, \theta_k)\right) \leq \sum^K_{k=1} P(\mathbf y\in I(\mathbf x, \theta_k)) =0.
\end{align*}
On the other hand, if $P(\mathbf y\in L(\mathbf x))>0$, instead of proving $P(\mathbf y\in \bar I(\mathbf x))=1$, it is equivalent to proving $P(\mathbf y\in \bar I^c(\mathbf x))=0$. Since $P(\mathbf y\in L(\mathbf x))= P(\mathbf y\in I(\mathbf x, \theta))>0$, there exists $0<p<1$, such that $P(\mathbf y\in I^c(\mathbf x,\theta)) = 1- P(\mathbf y\in I^c(\mathbf x, \theta))<p$. Thus, similar to the proof in \textbf{Theorem \ref{core-lemma}}
\begin{align*}
    P\left(\mathbf y\in \bar I^c(\mathbf x)\right) 
    &= P\left(\mathbf y\in\lim_{K\to\infty}\lim_{n\to\infty}\bigcup_{1\leq i_1<\cdots i_s\leq K\atop s>K-\sqrt{K}}\bigcap^s_{k=1} I^c(\mathbf x,\theta_{i_k})\right)\\
    &\leq \lim_{K\to\infty}\lim_{n\to\infty}\sum_{1\leq i_1<\cdots i_s\leq K\atop s>K-\sqrt{K}} P\left(\mathbf y\in\bigcap^s_{k=1} I^c(\mathbf x,\theta_{i_k})\right) \\
    &\leq \lim_{K\to\infty}\lim_{n\to\infty}\sum_{1\leq i_1<\cdots i_s\leq K\atop s>K-\sqrt{K}} p^s \\
    &\leq \lim_{K\to\infty}\lim_{n\to\infty}p^{K-\sqrt{K}}\left(\binom{K}{K}+\binom{K}{K-1}+\cdots \binom{K}{K-\left[\sqrt{K}\right]} \right)=0.
\end{align*}
After the discussion above, $P(\mathbf y\in L(\mathbf x))>0\Longrightarrow P(\mathbf y\in \bar I(\mathbf x))=1$, which completes the proof.
\end{proof}

\begin{proof}{\textbf{of Theorem \ref{well-define}}}\\
As $P(\mathbf y\in I(\mathbf x))<1$ and the zero-one law of LILI in \textbf{Corollary \ref{LILI-01}}, $P(\mathbf y\in I(\mathbf x))=0$, and using the equivalence (\ref{LI-equivalence}), we obtain $\lim_{n\to\infty}P(\mathbf y\in L(\mathbf x))<1$. Let $\mathbf w$ be an arbitrary instance. Since for every instance, the causal tree only divides it on one leaf node, $\mathbf w\in L(\mathbf x), \mathbf w\in L(\mathbf y)\Longrightarrow y\in L(\mathbf x)$. We have
\begin{align*}
    1 > \lim_{n\to\infty}P(\mathbf y\in L(\mathbf x)) 
     > \lim_{n\to\infty}P(\mathbf w \in L(\mathbf x); \mathbf w\in L(\mathbf y))
     > \lim_{n\to\infty}P(\mathbf w\in L(\mathbf x))P(\mathbf w\in L(\mathbf y)).
\end{align*}
Therefore, at least one of the probabilities $\lim_{n\to\infty}P(\mathbf w\in L(\mathbf x))$ and $\lim_{n\to\infty}P(\mathbf w\in L(\mathbf y))$ is less than one. Without losing generality, assume $P(\mathbf w\in L(\mathbf y))<1$, then $P(\mathbf w\in I(\mathbf y))=0$. Furthermore, $P(\mathbf w\in I(\mathbf x)\bigcap I(\mathbf y))\leq P(\mathbf w\in I(\mathbf y))=0$. As $\mathbf w$ is an arbitrary instance, 
\begin{align*}
    P\left(I(\mathbf x)\bigcap I(\mathbf y)\neq\emptyset\right)=0,
\end{align*}
and $P(I(\mathbf x)\bigcap I(\mathbf y)=\emptyset) = 1-P(I(\mathbf x)\bigcap I(\mathbf y)\neq\emptyset)=1$

Obviously, if $I(\mathbf x)=I(\mathbf y)$ on almost any $\Theta_\infty(\mathcal X_\infty)$, $\mathbf y\in I(\mathbf y)=I(\mathbf x)$ holds on the same almost any $\Theta_\infty(\mathcal X_\infty)$. It only needs to prove $P(\mathbf y\in I(\mathbf x))=1\Longrightarrow P(I(\mathbf x)=I(\mathbf y))=1$. For any $\mathbf w$ that $P(\mathbf w\in I(\mathbf x))=1$, 
\begin{align*}
    P(\mathbf w\in L(\mathbf y)) 
    \geq P(\mathbf w\in L(\mathbf x);\mathbf y\in L(\mathbf x))
    \geq P(\mathbf w\in L(\mathbf x))P(\mathbf y\in L(\mathbf x))
    \to 1.
\end{align*}
According to \textbf{Theorem \ref{core-lemma}}, $P(\mathbf w\in L(\mathbf y))\to 1\Longrightarrow P(\mathbf w\in I(\mathbf y))=1$, and $P(I(\mathbf x)\subset I(\mathbf y))=1$. We know $\mathbf y\in I(\mathbf x)\Longleftrightarrow \mathbf x\in I(\mathbf y)$. It is easy to infer that $P(I(\mathbf y)\subset I(\mathbf x))=1$ in a similar way, i.e., $P(I(\mathbf x)=I(\mathbf y))=1$. This equivalence has been proved.

\end{proof}

\begin{proof}{\textbf{of Theorem \ref{inter-convergence}}}\\
Without loss of generality, we set $p_\mathbf w = P(\mathbf w\in L(\mathbf x))$. According to the definition of $I(\mathbf x, \Theta_K)$ and discussion in \textbf{Section 3}, if $\mathbf w\in I(\mathbf x,\Theta_K)\cap I(\mathbf y,\Theta_K)$, there are at least $K-[\sqrt{K}]$ causal trees that classify $\mathbf x, \mathbf w$ or $\mathbf y, \mathbf w$ into the same leaves. These two sets of causal trees have overlap on at least $K-2[\sqrt{K}]$ causal trees. Therefore,
\begin{align*}
    \mathbf w\in I(\mathbf x,\Theta_K)\cap I(\mathbf y,\Theta_K)\subseteq \bigcup_{1\leq i_1<\cdots i_s\leq K\atop s> K-2\sqrt{K}}\bigcap_{k=1}^{s}(I(\mathbf x,\theta_{i_k})\cap I(\mathbf y,\theta_{i_k})).
\end{align*}
$\mathbf w\in I(\mathbf x,\theta_k)\cap I(\mathbf y, \theta_k)$ if and only if $\mathbf y\in L_k(\mathbf x)$ and $\mathbf w\in L_k(\mathbf x)$. By the randomness of $\mathbf w$, $P(\mathbf y\in L_k(\mathbf x))$ and $P(\mathbf w\in L_k(\mathbf x))$ are independent, and the distribution of $\mathbf w\in I(\mathbf x,\theta_k)\cap I(\mathbf y,\theta_y)$ follows binomial distribution $B(1,pp_\mathbf w)$. When considering all the $K$ causal trees, the binomial distribution becomes a 0-1 distribution $B(K,pp_\mathbf w)$.
\begin{align*}
    P(\mathbf w\in I(\mathbf x, \Theta_K)\cap I(\mathbf y, \Theta_K))
    & \leq P\left(\mathbf w\in\bigcup_{1\leq i_1<\cdots i_s\leq K\atop s> K-2\sqrt{K}}\bigcap_{k=1}^{s}(I(\mathbf x,\theta_{i_k})\cap I(\mathbf y,\theta_{i_k}))\right) \\
    & = \sum^{[2\sqrt{K}]}_{i=0}\binom{K}{i}(1-pp_\mathbf w)^i(pp_\mathbf w)^{K-[2\sqrt{K}]} \\
    & = (pp_\mathbf w)^K + \sum^{[2\sqrt{K}]}_{i=1}\binom{K}{i}(1-pp_\mathbf w)^i(pp_\mathbf w)^{K-[2\sqrt{K}]} \\
    & \leq (pp_\mathbf w)^K + (1-pp_\mathbf w)(pp_\mathbf w)^{K-[2\sqrt{K}]}\sum^{[2\sqrt{K}]}_{i=1}\binom{K}{i} \\
    & \leq (pp_\mathbf w)^K+ (1-pp_\mathbf w)(pp_\mathbf w)^{K-2\sqrt{K}}(2\sqrt{K})K^{2\sqrt{K}}.
\end{align*}
The same proof for the situation $p_\mathbf w = P(\mathbf w\in L(\mathbf y))$.

Similar to proof in \textbf{Theorem \ref{core-lemma}}, the upper bound of inequality (\ref{inter-bound}) converges to zero with $K,n\to\infty$. If $p_\mathbf w < 1$, 
\begin{align*}
\lim_{K\to\infty}\lim_{n\to\infty}\frac{P(\mathbf w\in I(\mathbf x,\Theta_K)\cap I(\mathbf y,\Theta_K))}{p^{K-2\sqrt{K}}} \leq\lim_{K\to\infty}\lim_{n\to\infty} p_\mathbf w^K+2(1-pp_\mathbf w)p_\mathbf w^{K-2\sqrt{K}}\sqrt{K}K^{2\sqrt{K}} = 0.
\end{align*}
By the assumption $p<1$, we directly obtain 
\begin{align*}
    \lim_{K\to\infty}\lim_{n\to\infty}\frac{P(\mathbf w\in I(\mathbf x,\Theta_K)\cap I(\mathbf y,\Theta_K))}{p_\mathbf w^{K-2\sqrt{K}}} \leq\lim_{K\to\infty}\lim_{n\to\infty} p^K+2(1-pp_\mathbf w) p^{K-2\sqrt{K}}\sqrt{K}K^{2\sqrt{K}} = 0.
\end{align*}
Combining the two equations above, the last part of the theorem can be proved.

\end{proof}

\begin{proof}{\textbf{of Lemma \ref{LILI-measure}}}\\
$I(\mathbf x, \theta_k, c+1)$ is formed by adding a split feature and a split value on $I(\mathbf x, \theta_k, c)$, if 
\begin{align*}
    i = \arg\min_j \frac{\mu(I(\mathbf x, \theta_k ,c+1)_j)}{\mu(I(\mathbf x, \theta_k, c)_j)},
\end{align*}
the extra split feature in $I(\mathbf x, \theta_k, c+1)$ is the $i$th feature since the measure on the other dimension remains the same, $\mu(I(\mathbf x, \theta_k, c+1)_j) = \mu(I(\mathbf x, \theta_k, c)_j)$ for $j\neq i$. If we only consider the discrete feature, the supremum of this ratio can be computed as 
\begin{align*}
    \eta(\mathbf X_{dis}) = \sup_{\mathbf x, \theta_k, c} \min_{X_j\in \mathbf X_{dis}}\frac{\mu(I(\mathbf x, \theta_k ,c+1)_j)}{\mu(I(\mathbf x, \theta_k, c)_j)} = 1-\frac{1}{C}<1
\end{align*}
where $\mathbf X_{dis}$ is the set of all discrete features, and $C$ represents the number of categories for the feature with the highest count of categories in $\mathbf X_{dis}$.
For all the continuous features $\mathbf X_{con}$, let's assume $\eta(\mathbf X_{con})=1$, then by the definition of supremum, there is a $\mathbf x$, $\theta_k$, $c$, and a split feature $i$ such that 
\begin{align*}
    \frac{\mu(I(\mathbf x, \theta_k, c+1)_i)}{\mu(I(\mathbf x, \theta_k, c)_i)} \geq 1-\delta,
\end{align*}
for any $\delta>0$. However, by the construction rules in \textit{$\alpha$-regularity}, the children nodes of split contain at least $l$ instances, i.e. there is $\mathbf x_1, ..., \mathbf x_l\in I(\mathbf x, \theta_k, c)/ I(\mathbf x, \theta_k, c+1)$. By the properties of the decision tree, $I(\mathbf x, \theta_k, c)_i$ is a continuous interval and can be represented as $[a_i^k, b_i^k]$. Since $\mu(I(\mathbf x, \theta_k, c)_i)=b^k_i-a^k_i$, without loss of generality, we set the split value as $(1-\delta)b$, and $I(\mathbf x, \theta_k, c+1)_i = [a_i^k, (1-\delta)b^k_i)$. Therefore, the values of instances $\mathbf x_1, ..., \mathbf x_l$ on the $i$th feature all belong to $[(1-\delta)b^k_i, b^k_i]$. Since $\delta$ can be any small enough number, there will always be $\mathbf x$, $\theta_k$, $c$ and feature $i$ that make the above equation hold, which means that we can find at least $l$ instances that have the same values on continuous feature $i$. This is contradictory to the assumption that there are no $l$ instances that have the same values on any continuous features. Thus, for all the continuous features $\eta(\mathbf X_{con})<1$, furthermore, 
\begin{align*}
    \eta = \max(\eta(\mathbf X_{dis}), \eta(\mathbf X_{con})) < 1.
\end{align*}
For the second part, if $\mathbf w, \mathbf y\in I(\mathbf x, \Theta_K)$, then there will be two sets of causal trees $\{\theta_{k_1}, ..., \theta_{k_m}\}$ and $\{\theta_{l_1}, ..., \theta_{l_m}\}$ that have $\mathbf w,\mathbf x$ and $\mathbf w, \mathbf y$ in the same leaves, where $m=K-[\sqrt{K}]$. Among these two sets of trees, there are at least $K-2[\sqrt{K}]$ causal trees that belong to both two sets. And on these overlap trees, $\mathbf x, \mathbf y,\mathbf w$ are in the same leaves. Hence, 
\begin{align*}
    P(\parallel\mathbf y-\mathbf w\parallel_1\geq \epsilon\mid \mathbf y,\mathbf w\in I(\mathbf x, \Theta_K)) 
    & \leq P\left(\parallel\mathbf y-\mathbf w\parallel_1\geq\epsilon\ \Bigg|\ \mathbf y,\mathbf w\in\bigcup_{1\leq i_1<\cdots i_s\leq K\atop s> K-2\sqrt{K}}\bigcap_{k=1}^{s}I(\mathbf x,\theta_{i_k})\right) \\
    & \leq P\left(\mu\left( \bigcap_{k=1}^{m}I(\mathbf x,\theta_{i_k})\right)\geq\epsilon\right).
\end{align*}
As the independent of each causal tree given dataset, we can set $\{\theta_{i_1}, ..., \theta_{i_m}\}$ as $\{\theta_1, ..., \theta_m\}$ without loss of generality. Using the chebyshev's theorem in the last step, we obtain 
\begin{align*}
P(\parallel\mathbf y-\mathbf w\parallel_1\geq \epsilon\mid \mathbf y,\mathbf w\in I(\mathbf x, \Theta_K)) \leq \frac{1}{\epsilon^2} E_\theta\left(\mu\left( \bigcap_{k=1}^{m}I(\mathbf x,\theta_{k})\right)\right).
\end{align*}
We set the boundary of the whole dataset as $\mathcal B = \prod_{i=1}^d[a_i, b_i]$, then by the definition of $\eta$, 
\begin{align*}
    \frac{\mu\left( \bigcap_{i=1}^{d}I(\mathbf x,\theta_{k})\right)}{\mu(\mathcal B)}
    \leq \frac{\sum^d_{i=1} \eta^{s_i}(b_i-a_i)}{\sum^m_{k=1}(b_i-a_i)}
    \leq \eta^{\min(s_1, ..., s_m)},
\end{align*}
where $s_i$ is the total number of times that $i$th feature is selected as a split feature along the road to $L(\mathbf x)$ in these $m$ causal trees. Furthermore, by the \textit{Random-split} rule, 
\begin{align*}
    E_\theta(s_i) \geq \frac{\pi}{d}\sum_{i=1}^dE_\theta(s_i) = \frac{\pi}{d}\sum^m_{k=1} E_\theta(D_k(\mathbf x)),
\end{align*}
where $D_k(\mathbf x)$ is the depth of $L(\mathbf x)$ in the $k$th causal tree and $E_\theta( D_K(\mathbf x))$ is the expected depth of leaf $L(\mathbf x)$ with respect to $\theta$. As the construction of each tree is independent, $E_\theta(D_1(\mathbf x))= \cdots E_\theta(D_K(\mathbf x)) = E_\theta(D(\mathbf x))$, and $E_\theta(s_i) \geq (\pi/d)mE_\theta(D(\mathbf x))$. Moreover, by the \textit{$\alpha$-regularity},
\begin{align*}
    n\alpha^{D(\mathbf x)} \geq l
    \Longrightarrow D(\mathbf x) \leq \frac{\ln n-\ln l}{-\ln\alpha}.
\end{align*}
Since the convex character of $\ln$, 
\begin{align*}
    \ln E_\theta\left(\eta^{\min(s_1, ..., s_m)}\right) 
    & \leq E_\theta(\min(s_1, ..., s_m)\ln\eta) \\
    & = \min(E_\theta(s_1), ... E_\mathcal D(s_m))\ln\eta \\
    & \leq \frac{\pi}{d} (K-\sqrt{K}) E_\theta(D(\mathbf x)) \ln\eta \\
    & \leq \frac{\pi}{d}(K-\sqrt{K})\frac{\ln n-\ln l}{-\ln\alpha}\ln\eta.
\end{align*}
As $\eta$ only depends on the dataset $\mathcal D$, it can be pulled out from expectation. Finally, 
\begin{align*}
E_\theta\left(\mu\left( \bigcap_{k=1}^{m}I(\mathbf x,\theta_{k})\right)\right)
\leq  E_\theta\left(\eta^{\min(s_1, ..., s_m)}\right)\mu(\mathcal B) \leq \eta^{\pi (K-\sqrt{K}) (\ln n-\ln l)/(-d\ln\alpha)}\mu(\mathcal B).
\end{align*}
Combining all the inequalities above, we finish our proof. 
\end{proof}

\begin{proof}{\textbf{Theorem \ref{LILI-size}}}\\
Recall the \textit{$\alpha$-regularity}, the child node of a split node at least contains $\alpha n$ samples, and the number of samples in a leaf node is between $l$ and $2l-1$. Hence, $l\leq\alpha n\leq 2l-1$, then 
\begin{align*}
    \frac{l}{n} \leq \lim_{n\to\infty} P(L(\mathbf x)) \leq \frac{2l-1}{n} \Longrightarrow \frac{\alpha}{2} < q_\mathbf x < 2\alpha.
\end{align*}

The probability that instances belong to $I(\mathbf x, \Theta)$ can be obtained as 
\begin{align*}
    P(I(\mathbf x, \Theta_K)) = \sum_\mathbf yP(\mathbf y\in I(\mathbf x, \Theta_K))P(\mathbf y).
\end{align*}
Like the proof in \textbf{Theorem \ref{core-lemma}}, 
\begin{align*}
    \lim_{n\to\infty} P(I(\mathbf x, \Theta_K)) 
    &\geq \lim_{n\to\infty} \sum_{\mathbf y}P\left(\mathbf y\in \bigcap_{k=1}^{K-[\sqrt{K}]} I(\mathbf x, \theta_k)\right)P(\mathbf y) \\
    & \geq \lim_{n\to\infty}\sum_\mathbf y\prod_{k=1}^{K-[\sqrt{K}]} P(\mathbf y\in I(\mathbf x, \theta_k))P(\mathbf y) \\
    & \geq \lim_{n\to\infty}\prod_{k=1}^{K-[\sqrt{K}]}\sum_\mathbf y P(\mathbf y\in I(\mathbf x, \theta_k))P(\mathbf y) \\
    & \geq \left(\frac{\alpha}{2}\right)^{K-[\sqrt{K}]}.
\end{align*}
Under the assumption $n = O\left((2/\alpha)^K\right)$, $|I(\mathbf x, \Theta_K)|$ goes to infinity in probability with respect to $n,K$ because 
\begin{align*}
    |I(\mathbf x, \Theta_K)| \overset{P}{\to} E_{\theta, \mathcal D}(|I(\mathbf x, \Theta_K)|) = nP(I(\mathbf x, \Theta))\geq n\left(\frac{\alpha}{2}\right)^{K-[\sqrt{K}]} \to \infty.
\end{align*}
\end{proof}

\begin{proof}{\textbf{of Theorem \ref{estimation-convergence}}}\\
It is well-known that if $Y\sim F(y\mid I(\mathbf x, \Theta_K))$, then $F(Y\mid I(\mathbf x, \Theta_K))$ follows a uniform distribution. Since $\mu(\mathcal B)<+\infty$, we can set the boundary of $Y$ as $[0,1]$ by norming the data, then $F(Y\mid I(\mathbf x, \Theta_K))\sim U([0,1])$. Let $U_i=F(Y_i\mid I(\mathbf X_i, \Theta_K))$,
\begin{align*}
    \widehat F(y\mid I(\mathbf x,\Theta_K)) 
    &= \sum^n_{i=1} W_i(I(\mathbf x, \Theta_K))\mathbf 1(Y_i\leq y) \\
    &= \sum^n_{i=1} W_i(I(\mathbf x, \Theta_K))\mathbf 1(U_i\leq F(y\mid I(\mathbf X_i, \Theta_K)))\\
    &= \sum^n_{i=1} W_i(I(\mathbf x, \Theta_K))\mathbf 1(U_i\leq F(y\mid I(\mathbf x, \Theta_K)) \\
    &+ \sum^n_{i=1} W_i(I(\mathbf x, \Theta_K))\left(\mathbf 1(U_i\leq F(y\mid I(\mathbf X_i, \Theta_K) - \mathbf 1(U_i\leq F(y\mid I(\mathbf x, \Theta_K)))\right).
\end{align*}
The second step is based on the \textit{monotonicity} assumption. As $F(y\mid \mathbf X_i)$ is strictly monotonously increasing, then $F(y\mid I(\mathbf X_i, \Theta_K))$ is also strictly monotonously increasing, and $Y_i\leq y \Longrightarrow F(Y_i\mid I(\mathbf X_i, \Theta_K)\leq F(y\mid I(\mathbf X_i, \Theta_K))$
Hence, the absolute difference between the true distribution and the estimated one can be transformed as 
\begin{align*}
    |F(y\mid I(\mathbf x, \Theta_K)) - &\widehat F(y\mid I(\mathbf x, \Theta_K))|
    \leq \big|F(y\mid I(\mathbf x, \Theta_K))-\sum^n_{i=1} W_i(I(\mathbf x, \Theta_K))\mathbf 1(U_i\leq F(y\mid I(\mathbf x,\Theta_K)))\big| \\
    &+ \big|\sum^n_{i=1} W_i(I(\mathbf x, \Theta_K))\mathbf 1(U_i\leq F(y\mid I(\mathbf X_i, \Theta_K))) - \mathbf 1(U_i\leq F(y\mid I(\mathbf x, \Theta_K)))\big|.
\end{align*}
For the first part, 
\begin{align}
\label{proof1}
    \big|F(y\mid I(\mathbf x, \Theta_K))-\sum^n_{i=1} W_i(I(\mathbf x, \Theta_K))\mathbf 1(U_i\leq F(y\mid I(\mathbf x,\Theta_K)))\big|
    &\leq \sup_{z\in [0,1]}\big|z-\sum^n_{i=1} W_i(I(\mathbf x, \Theta_K))\mathbf 1(U_i\leq z)\big|.
\end{align}
Let $\bar z$ be the value that makes the right side of this equation reach its supremum. For given $\mathbf x, \Theta_K, \mathcal D$, $\bar z$ and $W_i(I(\mathbf x, \Theta_K))$ can be treated as constant to $U_i$. Note that $E(\mathbf 1(U_i\leq \bar z)) = \bar z$, and 
\begin{align*}
    E\left(\sum^n_{i=1}W_i(I(\mathbf x, \Theta_K))\mathbf 1(U_i\leq \bar z)\right) = \bar z.
\end{align*}
Consequently, with the chebyshev’s theorem and the independence of $U_i$,
\begin{align*}
    &P\left(\big|F(y\mid I(\mathbf x, \Theta_K))-\sum^n_{i=1} W_i(I(\mathbf x, \Theta_K))\mathbf 1(U_i\leq F(y\mid I(\mathbf x,\Theta_K)))\big|\geq\epsilon\right)\\
    \leq & P\left(\big|\bar z-\sum^n_{i=1} W_i(I(\mathbf x, \Theta_K))\mathbf 1(U_i\leq \bar z)\big|\geq \epsilon\right) \\
    \leq & \frac{\bar z^2}{\epsilon^2} \sum^n_{i=1} W^2_i(I(\mathbf x, \Theta_K)).
\end{align*}
By the \textbf{Lemma \ref{LILI-size}}, $W_i(I(\mathbf x, \Theta_K)) \leq 1/|I(\mathbf x, \Theta_K)| \to 0$, then $W(I(\mathbf x, \Theta_K)) = \max_i(W_i(I(\mathbf x, \Theta_K))) \to 0$. Consequently, 
\begin{align*}
    \lim_{K\to\infty}\lim_{n\to\infty}\sum^n_{i=1} W_i^2(\mathbf x, \Theta_K)) \leq\lim_{K\to\infty}\lim_{n\to\infty} W(I(\mathbf x, \Theta_K))\sum^n_{i=1} W_i(I(\mathbf x, \Theta_K)) =\lim_{K\to\infty}\lim_{n\to\infty}W(I(\mathbf x, \Theta_K)) = 0.
\end{align*}
Hence, the first part of the absolute difference converges to 0 in probability,
\begin{align*}
    \big|F(y\mid I(\mathbf x, \Theta_K))-\sum^n_{i=1} W_i(I(\mathbf x, \Theta_K))\mathbf 1(U_i\leq F(y\mid I(\mathbf x,\Theta_K)))\big| \overset{P}{\to} 0.
\end{align*}
Note that in the second part in \ref{proof1},
\begin{align*}
    &E\left(\sum^n_{i=1} W_i(I(\mathbf x, \Theta_K))\mathbf 1(U_i\leq F(y\mid I(\mathbf X_i, \Theta_K))) - \mathbf 1(U_i\leq F(y\mid I(\mathbf x, \Theta_K)))\right) \\
    = & \sum^n_{i=1} W_i(I(\mathbf x, \Theta_K))\left(F(y\mid I(\mathbf X_i, \Theta_K)) - F(y\mid I(\mathbf x, \Theta_K))\right).
\end{align*}
Similar to the proof in the first part, we have 
\begin{align*}
    &\sum^n_{i=1} W_i(I(\mathbf x, \Theta_K))\mathbf 1(U_i\leq F(y\mid I(\mathbf X_i, \Theta_K))) - \mathbf 1(U_i\leq F(y\mid I(\mathbf x, \Theta_K))) \\
    \overset{P}{\to} & \sum^n_{i=1} W_i(I(\mathbf x, \Theta_K))\left(F(y\mid I(\mathbf X_i, \Theta_K)) - F(y\mid I(\mathbf x, \Theta_K))\right).
\end{align*}
Recall the definition of $W_i(I(\mathbf x, \Theta_K))$, which states that $W_i(I(\mathbf x, \Theta_K)) \neq 0$ if and only if the $i$th instance $\mathbf X_i\in I(\mathbf x, \Theta_K)$. It suffices to show that
\begin{align*}
    &\sum^n_{i=1} W_i(I(\mathbf x, \Theta_K))\left(F(y\mid I(\mathbf X_i, \Theta_K)) - F(y\mid I(\mathbf x, \Theta_K))\right) \\
  = & \sum^n_{\mathbf X_i\in I(\mathbf x, \Theta_K)} \frac{1}{|I(\mathbf x, \Theta_K)|} \left(F(y\mid I(\mathbf X_i, \Theta_K)\cap I(\mathbf x, \Theta_K)) - F(y\mid I(\mathbf x, \Theta_K))\right) \\
   \leq & \sup_{\mathbf x, \mathbf x' \in I(\mathbf x, \Theta_K)}\parallel\mathbf x-\mathbf x'\parallel_1 L.
\end{align*}
The last step is the direct result of \textit{Lipschitz continuity} assumption. Therefore, by the conclusion from \textbf{Lemma \ref{LILI-measure}}, 
\begin{align*}
    \sup_{\mathbf x, \mathbf x' \in I(\mathbf x, \Theta_K)}\parallel\mathbf x-\mathbf x'\parallel_1 L = L\mu\left( I(\mathbf x, \Theta_K)\right) \overset{P}{\to} 0
\end{align*}
which states the convergence in probability of the second part. 
\begin{align*}
    &\sum^n_{i=1} W_i(I(\mathbf x, \Theta_K))\mathbf 1(U_i\leq F(y\mid I(\mathbf X_i, \Theta_K))) - \mathbf 1(U_i\leq F(y\mid I(\mathbf x, \Theta_K))) \\
    \overset{P}{\to} & \sum^n_{i=1} W_i(I(\mathbf x, \Theta_K))\left(F(y\mid I(\mathbf X_i, \Theta_K)) - F(y\mid I(\mathbf x, \Theta_K))\right)
    \overset{P}{\to} 0.
\end{align*}

Combining the convergence of these two parts, we complete the proof of convergence on estimated distribution. Using the transformation in \ref{proof1}, we can infer that 
\begin{align*}
     &P(|\widehat F(y\mid I(\mathbf x, \Theta_K)) - F(y\mid I(\mathbf x, \Theta_K))|\geq \epsilon) \\
    \leq &\frac{1}{\epsilon^2}W(I(\mathbf x, \Theta_K)) + P\left(L\parallel\mathbf x-\mathbf x'\parallel_1\geq \epsilon \mid \mathbf x, \mathbf x'\in I(\mathbf x,\Theta_K)\right) \\
    \leq &\frac{1}{\epsilon^2nq_\mathbf x^K} + \frac{L^2}{\epsilon^2} \eta^{\pi K (\ln n-\ln l)/(-d\ln\alpha)}\mu(\mathcal B).
\end{align*}
Notice that $W_i(I(\mathbf x, \Theta_K))$ is zero or $\frac{1}{|I(\mathbf x, \Theta_K)|}$. Hence, $W(I(\mathbf x, \Theta_K)) = 1/|I(\mathbf x, \Theta_K)| \leq 1/(nq_\mathbf x^K)$ by \textbf{Lemma \ref{LILI-size}}. 
\end{proof}

\begin{proof}{\textbf{of Theorem \ref{ATE-convergence}}}\\
Firstly, we need to connect the estimated distribution and estimated expectation. Let $\mathbf y\sim F(y\mid T=1, I(\mathbf x, \Theta_K))$ be the ground truth distribution of label and $\hat y\sim \widehat F(\mathbf y\mid T=1, I(\mathbf x, \Theta_K))$ be the estimated one. If $y$ is a discrete variable, and $y_1, ..., y_p$ are all the possible values and $y_1<\cdots <y_p$, then

\begin{align*}
    & E(\hat y\mid T=1, I(\mathbf x, \Theta_K)) \\
    = &\sum^p_{j=1}y_j\widehat P(y_j\mid T=1, \mathbf X\in I(\mathbf x, \Theta_K)) \\
    = &\sum^p_{j=2}y_j\left[\widehat F(y_j\mid T=1, I(\mathbf x, \Theta_K)) - \widehat F(y_{j-1}\mid T=1, I(\mathbf x,\Theta_K))\right] + y_1\widehat F(y_1\mid T=1, I(\mathbf x, \Theta_K))
\end{align*}
\begin{align*}
    = &\sum^n_{i=1} W_i(I_t(\mathbf x, \Theta_K))\left[\sum^p_{j=2}y_j\mathbf 1\left(y_{j-1}<Y_i\leq y_j\right)+y_1\mathbf 1(Y_i\leq y_1) \right] 
    = \frac{1}{|I_t(\mathbf x, \Theta_K)|}\sum_{i\in I(\mathbf x, \Theta_K)}Y_i 
\end{align*}
where $W_i(I_t(\mathbf x, \Theta_K)) = \mathbf 1(\mathbf X_i\in I_t(\mathbf x, \Theta_K))/|I_t(\mathbf x,\Theta_K)| $. For the situation that $\mathbf y$ is a continuous variable, assume that $[y_a, y_b]$ is the boundary of the label and $y_a\leq y_1 <\cdots y_p \leq y_b$ with $y_{j+1}-y_j = y_j - y_{j-1}$, by the definition of integral,
\begin{align*}
    E(\hat y\mid T=1, I(\mathbf x, \Theta)) 
    = &\lim_{p\to\infty} \sum^{p-1}_{j=1} y_j \frac{\widehat F(y_{j+1}\mid T=1, I(\mathbf x, \Theta_K))-\widehat F(y_j\mid T=1, I(\mathbf x, \Theta_K))}{y_{j+1}-y_j}\times (y_{j+1} - y_j) \\
    = &\sum^n_{i=1} W_i(I_t(\mathbf x, \Theta_K))\left[\lim_{p\to\infty}\sum^{p-1}_{j=1}y_j\mathbf 1\left(y_{j-1}<Y_i\leq y_j\right)+y_1\mathbf 1(Y_i\leq y_1) \right] \\
    = &\frac{1}{|I_t(\mathbf x, \Theta_K)|}\sum_{i\in I_t(\mathbf x, \Theta_K)}Y_i.
\end{align*}
Moreover, according to the result of \textbf{Theorem \ref{estimation-convergence}}, we know that 
\begin{align*}
    |\widehat P(y\mid T=1,\mathbf X\in I(\mathbf x,\Theta_K)) - P(y\mid T=1, \mathbf X\in I(\mathbf x,\Theta_K))|\overset{P}{\to} 0.
\end{align*}
For the discrete variable condition, 
\begin{align*}
    &|E(y\mid T=1, I(\mathbf x,\Theta_K)) - E(\hat y\mid T=1, I(\mathbf x,\Theta_K))| \\
    \leq & \sum^p_{j=1} y_j|P(y_j\mid T=1, \mathbf X\in I(\mathbf x, \Theta_K)) - \widehat P(y_j\mid T=1, \mathbf X\in I(\mathbf x, \Theta_K))| \overset{P}{\to} 0.
\end{align*}
The proof for the continuous variable condition is similar to the proof in estimated expectation. The same relationships hold for $I_c(\mathbf x, \Theta_K)$. In conclusion, 
\begin{align*}
    &|\widehat{ATE}(I(\mathbf x,\Theta_K)) - CATE| \\
    \leq & \bigg|\frac{1}{|I_t(\mathbf x, \Theta_K)|}\sum_{i\in I_t(\mathbf x, \Theta_K)} Y_i - E(y\mid T=1, I(\mathbf x, \Theta_K))\bigg|\\
    + & \bigg|\frac{1}{|I_c(\mathbf x, \Theta_K)|}\sum_{i\in I_c(\mathbf x, \Theta_K)} Y_i - E(y\mid T=0, I(\mathbf x, \Theta_K))\bigg| \overset{P}{\to} 0.
\end{align*}
By the CATE in (\ref{CATE}), the ground truth can be represented as 
\begin{align*}
    ATE = \sum_{I(\mathbf x, \Theta_K)\in \mathcal I}CATE(I(\mathbf x, \Theta_K))P(I(\mathbf x, \Theta_K)),
\end{align*}
where $P(I(\mathbf x, \Theta_K))$ is the true probability that samples fall into $I(\mathbf x, \Theta_K)$. Recall the estimated one 
\begin{align*}
    \widehat P(I(\mathbf x, \Theta_K)) = \frac{|I(\mathbf x, \Theta_K)|}{n} = \frac{1}{n}\sum^n_{i=1}\mathbf 1(\mathbf X_i\in I(\mathbf x, \Theta_K)).
\end{align*}
Therefore, $E(\widehat P(I(\mathbf x, \Theta_K))) = P(I(\mathbf x, \Theta_K))$. By the assumption, let $y\in [y_{\min}, y_{\max}]$. According to the definition, it is easy to derive that $|CATE(I(\mathbf x,\Theta_K))|, |\widehat ATE(I(\mathbf x, \Theta_K))|\leq 2y_{\max}$. Then
\begin{align*}
    |\widehat{ATE}-ATE| 
    & = \bigg|\sum_{I(\mathbf x, \Theta_K)\in \mathcal I} \widehat{ATE}(I(\mathbf x, \Theta_K))\widehat P(I(\mathbf x, \Theta_K)) - \sum_{I(\mathbf x, \Theta_K)\in\mathcal I} CATE(I(\mathbf x, \Theta_K)) P(I(\mathbf x, \Theta_K))\bigg| \\
    & \leq \bigg|\sum_{I(\mathbf x)\in\mathcal I}\left[\widehat{ATE}(I(\mathbf x, \Theta_K)) - CATE(I(\mathbf x, \Theta_K))\right]\widehat P(I(\mathbf x, \Theta_K))\bigg| \\
    & + \bigg|\sum_{I(\mathbf x, \Theta_K)\in\mathcal I}CATE(I(\mathbf x, \Theta_K))\left[ \widehat P(I(\mathbf x, \Theta_K)) - P(I(\mathbf x, \Theta_K)) \right]\bigg| \\
    & \leq \max_{I(\mathbf x, \Theta_K)\in \mathcal I}\left[\widehat{ATE}(I(\mathbf x, \Theta_K)) - CATE(I(\mathbf x, \Theta_K))\right] + 2y_{\max}\sum_{I(\mathbf x, \Theta_K)\in \mathcal I}\big|\widehat P(I(\mathbf x, \Theta_K)) - P(I(\mathbf x, \Theta_K))\big|.
\end{align*}
Since for any $I(\mathbf x, \Theta_K)$, $\widehat{ATE}(I(\mathbf x, \Theta_K))\overset{P}{\to} CATE(I(\mathbf x,\Theta_K))$, the first part also converges to zero in probability. For the second part, using the Hoeffding inequality, we can obtain
\begin{align*}
     P\left(\sum_{I(\mathbf x, \Theta_K)\in \mathcal I} \big|\widehat P(I(\mathbf x, \Theta_K)) - P(I(\mathbf x, \Theta_K))\big|\geq\epsilon\right) 
    &\leq  \sum_{I(\mathbf x, \Theta_K)\in \mathcal I}P\left( \big|\widehat P(I(\mathbf x, \Theta_K)) - P(I(\mathbf x, \Theta_K))\big|\geq\epsilon \right) \\
    & \leq |\mathcal I|\exp(-2n^2\epsilon^2).
\end{align*}
As we analyze in \textbf{Section 4}, $|\mathcal I|\leq \frac{1}{q_\mathbf x^K}$ and $n = O\left(\frac{1}{q_\mathbf x^K}\right)$ for any $\mathbf x$, 
\begin{align*}
    \lim_{K\to\infty}\lim_{n\to\infty} |\mathcal I|\exp(-2n^2\epsilon^2) = 0.
\end{align*}
Therefore, combining these two parts, $\widehat{ATE}\overset{P}{\to} ATE$.
\end{proof}

\begin{proof}{\textbf{of Theorem \ref{LILI-extention1}}}\\
According to the proof of \textbf{Theorem \ref{core-lemma}}, we have the upper bound of LILI under $f$ as 
\begin{align*}
    P(\mathbf y\in I(\mathbf x, f)) \leq \lim_{K\to\infty}p^{K-f(K)}\left(\binom{K}{K}+\binom{K}{K-1}+\cdots \binom{K}{K-[f(K)]} \right).
\end{align*}
Using different methods of magnification, we can obtain the first two parts of this theorem. Firstly, if $p<1$ and $f$ satisfies inequality (\ref{f-require1}), then
\begin{align*}
    P(\mathbf y\in I(\mathbf x,f)) \leq \lim_{K\to\infty}p^{K-f(K)} (1+K^{f(K)+1})
\end{align*}
Considering $\ln\left(p^{K-f(k)} K^{f(K)+1}\right) = (K-f(K))\ln p + (f(K)+1)\ln K$, by inequality (\ref{f-require1}), $\lim_{K\to\infty}f(K)/K = 0$, and 

\begin{align*}
    \lim_{K\to\infty}\frac{(K-f(K))\ln p+(f(K)+1)\ln K}{K\ln p} = 1 + \lim_{K\to\infty}\frac{f(K)\ln K}{K\ln p} > 2,
\end{align*}
\begin{align*}
    p^{K-f(K)}K^{f(K)+1} =\exp\left((K-f(K))\ln p+(f(K)+1)\ln K\right) \leq \exp(2K\ln p)\to 0.
\end{align*}
Thus, $P(\mathbf y\in I(\mathbf x,f))\to 0$, which proves $\lim_{n\to\infty} L(\mathbf y\in L(\mathbf x)) <1 \Longrightarrow P(\mathbf y\in I(\mathbf x,f))=1$. For the second part, as $p<1/2$, 
\begin{align*}
    P(\mathbf y\in I(\mathbf x, f)) \leq \lim_{K\to\infty}p^{K-f(K)}\sum^K_{i=1} \binom{K}{i} = \lim_{K\to\infty}p^{K-f(K)}2^K = 0.
\end{align*}
The proof of the last part is the same as \textbf{Theorem \ref{core-lemma}}, i.e.,
\begin{align*}
    P(\mathbf y\in I(\mathbf x))
    \geq P\left(\mathbf y\in\lim_{K\to\infty}\lim_{n\to\infty}\bigcap_{k=1}^K I(\mathbf x, \theta_k)\right)
    = \lim_{K\to\infty}\lim_{n\to\infty}\prod^K_{k=1}P(\mathbf y\in I(\mathbf x, \theta_k)) = 1.
\end{align*}
As this process has no assumption for $f$, then $\lim_{n\to\infty}P(\mathbf y\in L(\mathbf x)) = 1\Longrightarrow P(\mathbf y\in I(\mathbf x)) = 1$ holds for any $f$.
\end{proof}

\begin{proof}{\textbf{of Theorem \ref{LILI-extention2}}}\\
Recall the definition of $I(\mathbf x, K, f)$. If we divide $K$ trees into $K/(K-[f(K)])$ groups, then 
\begin{align*}
    \bigcup^{[K/m]-1}_{s=0} I_s = \bigcup^{[K/m]-1}_{s=0}\bigcap^{m+s*m}_{k=1+s*m} I(\mathbf x, \theta_k) \subset I(\mathbf x, \Theta_K, f)
\end{align*}
where $m = K- [f(K)]$. Since each $I_s$ is consisted of different trees, and the leaves of different causal trees $I(\mathbf x, \theta_k), k=1,2,..., K$ are independent of each other. Thus, $I_s, s= 1,2,..., [K/m]-1$ are also independent of each other. We have
\begin{align*}
    \lim_{n\to\infty}P(\mathbf y\in I(\mathbf x, \Theta_K, f)) &\geq\lim_{n\to\infty} P\left(\mathbf y\in \bigcup^{[K/m]-1}_{s=0} I_s\right) \\
    & = 1-\lim_{n\to\infty}\prod^{[K/m]-1}_{s=0}P\left(\mathbf y \notin I_s\right) \\
    & = 1 - (1-p^m)^{[K/m]-1}.
\end{align*}
The last two steps are based on the equation $P(\cup_i A_i) = 1 - \prod_iP\left(\overline{A_i}\right)$. If $m = K - [f(K)] \to c$ where $c$ is a constant, then $1-p^m < 1$, $K/m\to \infty$ and
\begin{align*}
    \lim_{K\to\infty}(1-p^m)^{[K/m]- 1} = 0.
\end{align*}
Otherwise, $m\to\infty$, the limit becomes
\begin{align*}
    \lim_{K\to\infty}(1-p^m)^{K/m} = \lim_{K\to\infty}\left((1-p^m)^{\frac{1}{p^m}}\right)^{p^mK/m} = \lim_{K\to\infty}\exp\left(-\frac{K}{m}p^m\right).
\end{align*}
According to the assumption to $f$ in (\ref{f-require2}), $\ln m/\ln K\to 0$. Then let us consider
\begin{align*}
    \lim_{K\to\infty}\frac{\ln\left(\frac{K}{m}p^m\right)}{\ln K} &= \lim_{n\to\infty} 1 - \ln p\frac{K-f(K)}{\ln K} = 1,\\
    \lim_{K\to\infty}\ln\left(\frac{K}{m}p^m\right) &= \lim_{K\to\infty}\ln K = +\infty, \\
    \lim_{K\to\infty}(1-p^m)^{K/m} &= \lim_{K\to\infty}(1-p^m)^{[K/m]-1} = 0.
\end{align*}
Therefore, for both conditions, $P(\mathbf y\in I(\mathbf x, \Theta_K, f))\to 1$. Furthermore, if we know $p > 0.5$, and $f$ satisfies (\ref{f-require3})
\begin{align*}
    \lim_{K\to\infty}(1-p^m)^{K/m} = \lim_{K\to\infty}\exp\left(-\frac{K}{m}p^m\right) \leq \lim_{K\to\infty} \exp\left(-\frac{K}{K-f(K)}\left(\frac{1}{2}\right)^{K-f(K)}\right) = 0.
\end{align*}
Hence, $p>0.5\Longrightarrow P(\mathbf y\in I(\mathbf x, \Theta_K, f))=1$.
\end{proof}

\section{}
This part provides the numerical results of Figure \ref{TWINS-ATE}, Figure \ref{bar-IHDP}, and Figure \ref{TWINs-ITE}.
\begin{table*}[htpb]
\centering
\fontsize{10}{15}\selectfont
\caption{$L_1$ loss of ATE on TWINS datasets}
\label{L1-TWINS}
\setlength{\tabcolsep}{3mm}{
\begin{tabular}{cccc}
\bottomrule[1.5pt]
\hline
\multirow{2}{*}{Algorithms} & \multirow{2}{*}{gap=500} & \multirow{2}{*}{gap=700} & \multirow{2}{*}{gap=900} \cr
 & \cr
\bottomrule
S-learner     & 0.0058(0.001) & 0.0053(0.001) & 0.0057(0.001)  \cr
T-learner     & 0.0030(0.001) & 0.0104(0.001) & 0.0222(0.001)  \cr
R-learner     & 0.0042(0.001) & 0.0094(0.001) & 0.0223(0.003)  \cr
DML           & 0.0046(0.001) & 0.0029(0.002) & 0.0139(0.003)  \cr
Causal Forest & 0.0027(0.002) & 0.0013(0.001) & 0.0065(0.001)  \cr
Uplift Forest & 0.0023(0.003) & 0.0020(0.001) & 0.0097(0.002)  \cr
IPW           & 0.0038(0.000) & 0.0014(0.000) & 0.0080(0.000)  \cr
DRIV-learner  & 0.0047(0.001) & 0.0039(0.002) & 0.0297(0.003)  \cr
TMLE          & 0.0032(0.002) & 0.0018(0.001) & 0.0154(0.006)  \cr
Optim         & 0.0049(0.000) & 0.0051(0.000) & 0.0098(0.000)  \cr
RNNM           & 0.0024(0.000) & 0.0022(0.000) & 0.0094(0.000)  \cr
LILI          & \textbf{0.0017(0.001)} & \textbf{0.0005(0.002)} & \textbf{0.0007(0.001)} \cr
 \bottomrule[1.5pt]
\end{tabular}}
\end{table*}

\begin{table*}[htpb]
\centering
\fontsize{10}{13}\selectfont
\caption{PEHE of ITE on IHDP datasets}
\label{PEHE-IHDP}
\setlength{\tabcolsep}{3mm}{
\begin{tabular}{cccccc}
\bottomrule[1.5pt]
\hline
\multirow{2}{*}{Algorithms} & \multirow{2}{*}{IHDP1} & \multirow{2}{*}{IHDP2} & \multirow{2}{*}{IHDP3} & \multirow{2}{*}{IHDP4} & \multirow{2}{*}{IHDP5}  \cr
 & \cr
\bottomrule
S-learner      & 14.32(0.179) & 60.07(0.009) & 73.26(0.249) & 55.65(0.016) & 40.53(0.13) \cr
T-learner      & 15.34(0.299) & 60.0(0.028) & 73.3(0.167) & 55.56(0.249) & 49.88(0.871) \cr
R-learner      & 12.54(0.238) & 40.04(1.013) & 47.48(2.087) & 36.46(0.478) & 39.93(0.742) \cr
DML            & 54.94(11.288) & 66.91(2.923) & 80.93(2.138) & 63.32(4.053) & 80.38(26.422) \cr
Causal Forest  & 12.33(0.130) & 59.07(0.134) & 66.78(0.179) & 54.45(0.036) & 40.22(0.462) \cr
Uplift Forest  & 24.95(0.058) & 53.25(0.091) & 90.1(0.16) & 31.65(0.063) & 57.15(0.417) \cr
IPW            & 31.78(0.000) & 14.02(0.000) & 53.97(0.000) & 11.8(0.000) & 70.34(0.000) \cr
DRIV-learner   & 14.53(0.613) & 56.25(2.16) & 72.05(1.454) & 52.39(1.199) & 48.01(0.827) \cr
TMLE           & 12.43(0.209) & 59.5(0.336) & 67.77(0.079) & 54.64(0.058) & 42.03(0.669) \cr
Optim          & 10.55(0.000) & 56.95(0.000) & 67.45(0.000) & 54.73(0.000) & 30.88(0.000) \cr
RNNM            & \textbf{10.33(0.000)} & 56.16(0.000) & 66.15(0.000) & 54.23(0.000) & 29.89(0.000) \cr
LILI           & 10.94(0.084) & \textbf{13.3(0.021)} & \textbf{14.67(0.099)} & \textbf{13.52(0.039)} & \textbf{29.31(0.154)} \cr
 \bottomrule[1.5pt]
\end{tabular}}
\end{table*}

\begin{table*}[htpb]
\centering
\fontsize{10}{15}\selectfont
\caption{PEHE of ITE on TWINS datasets}
\label{PEHE-TWINS}
\setlength{\tabcolsep}{3mm}{
\begin{tabular}{cccc}
\bottomrule[1.5pt]
\hline
\multirow{2}{*}{Algorithms} & \multirow{2}{*}{gap=500} & \multirow{2}{*}{gap=700} & \multirow{2}{*}{gap=900} \cr
 & \cr
\bottomrule
S-learner     & 0.0260(0.0003) & 0.0332(0.0001) & 0.0434(0.0006)  \cr
T-learner     & 0.0203(0.0001) & 0.0305(0.0002) & 0.0415(0.0007)  \cr
R-learner     & 0.0233(0.0007) & 0.0364(0.0007) & 0.0490(0.0013)  \cr
DML           & 0.0524(0.0003) & 0.0761(0.0027) & 0.1333(0.0096)  \cr
Causal Forest & 0.0298(0.0001) & 0.0420(0.0001) & 0.0563(0.0001)  \cr
Uplift Forest & 0.0295(0.0001) & 0.0416(0.0001) & 0.0561(0.0003)  \cr
IPW           & 0.0281(0.0000) & 0.0351(0.0000) & 0.0346(0.0000)  \cr
DRIV-learner  & 0.0303(0.0001) & 0.0439(0.0001) & 0.0681(0.0018)  \cr
TMLE          & 0.0232(0.0001) & 0.0342(0.0002) & 0.0478(0.0005)  \cr
Optim         & 0.0303(0.0000) & 0.0428(0.0000) & 0.0572(0.0000)  \cr
RNNM           & 0.0303(0.0000) & 0.0427(0.0000) & 0.0572(0.0000)  \cr
LILI          & \textbf{0.0170(0.0010)} & \textbf{0.0258(0.0001)} & \textbf{0.0284(0.0002)}  \cr
 \bottomrule[1.5pt]
\end{tabular}}
\end{table*}

\vskip 0.2in
\bibliography{sample}

\end{document}